\documentclass{article}
\usepackage[preprint]{neurips_2026}
\usepackage{hyperref}
\usepackage{url}
\usepackage{amsmath, amssymb, amsthm}
\usepackage{graphicx}
\usepackage{booktabs}
\usepackage{cleveref}
\usepackage{natbib}
\usepackage{subcaption}
\usepackage[section]{placeins}

\usepackage{algorithm}
\usepackage{algpseudocode}

\usepackage{float}
\title{Learning Holographic Reduced Representations with Clifford Variational Autoencoders}

\author{
  Mohamed Malek Abid \\
  Institute of Neuroinformatics, \\
  University of Zurich \& ETH
Zurich\\
  \texttt{moabid@ethz.ch}\thanks{Corresponding author: \texttt{moabid@ethz.ch}.} \\
  \And
  P. Michael Furlong \\
  National Research Council Canada \& \\
  Centre for Theoretical Neuroscience, \\
  Systems Design Engineering, \\
  University of Waterloo, Waterloo, Canada\\
  \texttt{michael.furlong@nrc-cnrc.gc.ca} \\
}

\begin{document}

\maketitle

\begin{abstract}

Vector Symbolic Algebras project data structures into a hyperdimensional vector space through the application of their vector algebras to randomly generated atomic vector symbols and fractional power encodings of real-valued data. Embedding unstructured data remains an open question. We present \textit{Clifford-VAE}, a variational autoencoder that learns to project data onto a Clifford torus in arbitrary dimensions.  
Experiments using the MNIST, FashionMNIST, and CIFAR-10 datasets demonstrate that Clifford-VAE produces representations that are competitive with those produced by Gaussian and Hyperspherical VAEs for semi-supervised classification tasks while outperforming Gaussian and Hyperspherical counterparts in the VSA benchmark tests of self-binding and unbinding, role-filler recovery, and bundle capacity. Clifford-VAE provides a principled technique for grounding perceptual data into a symbolic reasoning framework, providing a new approach to a long-standing problem in the VSA literature.
\end{abstract}

%%%%%%%%%%%%%%%%%%%%%%%%%%%%%%%%%%%%%%%%%%%%%%%%%%%%%%%
\section{Introduction}
% oy distributed representations that are randomly sampled that, with sufficiently high dimensionality, demonstrate pseudo-orthogonality and permit symbol-like reasoning in neural networks.

Vector Symbolic Algebras (or Architectures~\citep{Levy2008VectorSA}) represent symbolic information using high-dimensional
distributed vectors, enabling neuro-symbolic computation through algebraic operations on
these representations~\citep{gayler-2004}. VSAs enable the manipulation of vectors according to an algebra defined over symbols that are identified with randomly generated \emph{atomic} vectors. VSAs can also incorporate real-valued vector data through \emph{fractional binding} or \emph{fractional power encoding}~\citep{plate1992,komer2019,frady-2021}, as well as through \emph{recursive} applications of fractional binding~\citep{simulating-voelker-2021}. Structured data can then be embedded in a VSA representation through the application of its algebra. However, there remains an open question of the best method to embed data into a VSA representation, something particularly challenging for unstructured data. We address this problem for the Holographic Reduced Representations~\citep{plate_holographic_1995,plate2003holographic} algebra.

Canonically, HRR vectors are generated by normalizing
$\mathbf{x} \sim \mathcal{N}(\mathbf{0}, \mathbf{I}/d)$. However, generating vectors via the method:
\begin{equation}
\label{eq:unitary_sampling}
    \mathbf{x} = \mathcal{F}^{-1}\left\{e^{i\mathbf{a}}\right\};\quad
    (a_{i}) \sim \mathcal{U}[-\pi,\pi]
\end{equation}
ensures all Fourier coefficients have unit magnitude,
$|\mathcal{F}\{\mathbf{x}\}[k]| = 1\;\forall k$, making binding exactly invertible in the
Fourier domain. 
Vectors of this form are characterized as
  \textit{circular} HRRs (later FHRRs) in \citet[Ch.~4]{plate2003holographic}, or \emph{unitary} (in the real domain)~\citep{komer2019},
  where each element is represented directly as the phase of a frequency
  component, enabling binding in 
  $O(d)$ time if using the FHRR (Fourier Holographic Reduced Representation) schema.
Unitarity is also a necessary property for a vector to be fractionally bound~\citep{komer2019}. Fractional binding occurs when a base vector is bound
  with itself a real-valued number of times by exponentiation in the Fourier domain,
  $\mathbf{b}^k = \mathcal{F}^{-1}\{\mathcal{F}\{\mathbf{b}\}^k\}$ for $k \in \mathbb{R}$.
In the non-unitary case, the binding operation's approximate inverse experiences degradation,
often necessitating the use of clean-up memories~\citep{plate_holographic_1995}, an important
factor of the computational cost involved with practical HRR-based solutions for tasks such
as audio fingerprinting~\citep{fujita2024audiofingerprintingholographicreduced}. This data embedding technique has been used in spiking models of spatial memory and SLAM as well as compressed, queryable image representations~\citep{dumont2023b, penzkofer25_aic}.

% For details, see \Cref{appendix:hrr_math}.

%Two fundamental operations involved in a VSA are superposition, also called bundling, and a form of invertible composition referred to as \textit{binding}~\citep{smolensky1990}. In the Holographic Reduced Representation VSA, these \textit{atomic} vectors are traditionally initialized through random sampling, which, provided the dimensionality is sufficiently high, yields pseudo-orthogonal \textit{distributed representations} which enable symbol-like reasoning in neural networks. In such an instantiation, atomic vectors are sampled randomly and identified with discrete objects, thus enabling symbolic reasoning can then be implemented through the VSA's algebraic operations. Random sampling is the default mechanism for producing atomic vectors in the HRR algebra~\citep{plate1992}. 

In this paper we present Clifford-VAE, an unsupervised method for embedding unstructured data into our algebra of choice, Holographic Reduced Representations. We achieve this by applying a prior on the embedding layer of a variational autoencoder that forces latent representations to lie on a hyperdimensional Clifford torus. We also impose the constraint that latent representations have conjugate symmetry, thus ensuring the sampled complex representations can be converted to real-valued representations through the inverse Fourier transform. This results in producing vectors that have been previously referred to as \textit{unitary} vectors. These embeddings are characterized by having Fourier components with unit magnitude.

VSAs are compatible with vector representations learned from data, but the learning process requires additional training time and data. Deep learning approaches to generate HRR vectors have been explored, albeit only in the context of improving performance for supervised learning using a superposition of `feature descriptors' from intermediate layers of a neural network rather than learning the representation itself~\citep{Neubert2021}.

An early effort to learn \textit{reduced representations} was undertaken
by \citet{pollack1991}, who framed the problem under attack as ``representing variable-sized symbolic sequences or trees in a numeric fixed-width format''. In his work on 
Recursive Auto-associative Memories (RAAMs), the specific representation is not constrained to a VSA framework such as the HRR; rather, a \textit{reconstructor} and a \textit{compressor} are trained (via backpropagation) to decompose a hierarchy of distributed representations in a pre-specified structure, \textit{e.g.} a binary tree with fixed leaves. This approach mirrors ours with respect to the fact that we are able to encode and decode unstructured data to and from variable-length structures with the Clifford-VAE. Additionally, \citet{pollack1991} raised concerns regarding the limitations of symbolic systems, specifically, that they operate on information-free atoms which must be constrained (or \textit{cleaned-up}) to the set of \textit{semantically interpretable}
atoms after computations. Thus, our work can be considered to be an extension of 
this line of inquiry, as the Clifford-VAE enables the construction of symbols that are grounded in perceptual data and, potentially, a system that combines those symbols \textit{only} 
in a systematic fashion.

Subsequent efforts at the intersection of learning and Holographic Reduced Representations have primarily focused on task-specific, VSA-based augmentations of supervised learning approaches.
\citet{eliasmith2005} showed the ability of simple models to learn abstract rules when learning is constrained by the HRR algebra. \citet{ganesan2021learningholographicreducedrepresentations}
trained through HRR operations for extreme multi-label classification by normalizing Fourier
coefficients of intermediate representations, but froze randomly initialized vectors during
training; allowing Gaussian-sampled representations to be learned degraded classification
performance. This work was extended to circular vectors by \citet{nishida-etal-2024-multi}. \citet{nickel2016}
used circular convolutions to create more powerful knowledge graph embeddings, evaluating on discriminative tasks that did not
require preserving invertibility. \citet{alam2023recastingselfattentionholographicreduced} evaluated a superimposed memory with circularly convolved key-value vector pairs as a linear-cost substitute for attention, yielding state-of-the-art performance on sequences longer than $T=100,000$. Finally, Spatial Semantic Pointers, a form of fractional power encoding, have been used as a powerful and flexible image encoding schema that enables sample efficient learning on downstream tasks through the continuous fixed-input distributed representations that SSPs provide~\citep{penzkofer25_aic}. The related concept of learned Fourier features~\citep{li2021functional} has found embeddings of data that are inherently compatible with the (F)HRR algebra, although they have not been presented in that context, and are typically employed in supervised learning tasks.

In order to provide unsupervised learning for HRR-compatible embeddings, we turn to Variational Autoencoders~\citep[VAEs;][]{kingma2013}. VAEs combine probabilistic modeling with deep learning by learning a latent variable model
$p(\mathbf{x}|\mathbf{z})$ parameterized by neural networks~\citep{kingma2013}. The encoder
network approximates the posterior $p(\mathbf{z}|\mathbf{x})$ with a variational
distribution $q_\phi(\mathbf{z}|\mathbf{x})$, while the decoder represents the likelihood
$p_\theta(\mathbf{x}|\mathbf{z})$. Training optimizes the Evidence Lower Bound (ELBO):
\begin{equation}
\mathcal{L}(\theta, \phi; \mathbf{x}) =
  \mathbb{E}_{q_\phi(\mathbf{z}|\mathbf{x})}[\log p_\theta(\mathbf{x}|\mathbf{z})]
  - D_{KL}(q_\phi(\mathbf{z}|\mathbf{x}) \| p(\mathbf{z})).
\end{equation}
However, to provide latent representations that are compatible with the HRR algebra, we need representations that are at least embedded on the hypersphere, and ideally unitary.

The unit hypersphere has been shown to be a powerful feature space across representation learning. \citet{pmlr-v119-wang20k} showed that contrastive loss tends to align data on the hypersphere, and that loss metrics that directly reward alignment between positive examples \emph{and} uniformity of the induced distribution on the hypersphere improves the utility of learned representations in other tasks, relative to simple contrastive learning alone. This suggests that the hypersphere is the ``right'' space to embed data. L2-normalized latents have also recently proven themselves in generative modeling at scale, where constraining autoregressive image models to hyperspherical latent tokens was shown to remove the issues responsible for variance collapse during decoding and yielded state-of-the-art FID for AR methods on ImageNet~\citep{ke2026hyperspherical}.
% \subsection{Hyperspherical Variational Autoencoders}

Prior work on VAEs with hyperspherical priors is somewhat limited. \citet{davidson2018} replaced the Gaussian prior with a von Mises-Fisher (vMF) distribution
on $\mathcal{S}^{d-1}$; \citet{decao2020} introduced the Power Spherical distribution as a
more tractable alternative. While both enforce unit norm $\|\mathbf{z}\| = 1$, neither
constrains individual Fourier coefficient magnitudes. Both suffer from surface-area
concentration at high $d$: posterior mass collapses toward the equator relative to the mean
direction~\citep{xu2018}.

% \subsubsection{Comparison to Product-Space Hyperspherical VAEs}

\citet{increasingexpressivitydavidson} partially addressed this by decomposing a single
hypersphere into a product of smaller spheres $\mathcal{S}^{a_1} \times \cdots \times
\mathcal{S}^{a_k}$; recently, \citet{sablica2025spcauchy} proposed the spherical Cauchy
distribution as a numerically stable alternative, and improving 
sampling methods for the von Mises-Fisher is still an active area of research \citep{kim-2021}.
We instead
construct $\mathcal{CT}^{d-1} = (\mathcal{S}^1)^{d-1}$, the maximal factorization where
every factor is a circle. This specific choice is what enforces per-component unit Fourier
magnitudes; any factor with $a_i > 1$ would permit magnitude variation within that factor's
subspace and break exact HRR invertibility.
As our Clifford torus is the special case
where each factor is $\mathcal{S}^1$, we can sample from $\mathcal{S}^1$ with a von Mises/von Mises-Fisher implementation or, alternatively, \citet{decao2020}'s Power Spherical distribution can be utilized. 

Our approach is motivated by two concerns: first, the notion that while randomly initializing high-dimensional vectors guarantees that vectors representing distinct objects neatly integrate into existing algebraic methods, they retain no grounding in perceptual data. Thus, a 
cognitive map built with randomly initialized SSPs can answer where a landmark is, but it holds no perceptual account of the landmark itself, at least not implicitly. Second, current methods require intuiting embeddings or relying on pre-trained features, and do not take into account the constraints of embedding data in unitary vectors when learning features.

To address these problems we present the Clifford Variational Autoencoder (Clifford-VAE), a variational account for learning feature embeddings in hyperspherical spaces that are compatible with the HRR/FHRR vector symbolic algebras. We achieve this by replacing the standard isotropic Gaussian distribution over the hidden layer of a variational autoencoder with a series of independent von Mises distributions and enforcing conjugate symmetry on these representations before applying a discrete Fourier transform.

In our work, we train with an objective that combines reconstruction (aligning each latent with its input), and uniformity, due to our KL term against the uniform prior on $\mathcal{CT}^{d-1}=(\mathcal{S}^1)^{d-1}$. However, given our construction, we also enforce the additional constraint that is provided by our Clifford torus parameterization, namely, preserved invertibility under circular convolution, expanding on the work of \citet{increasingexpressivitydavidson}, and integrating with the VSA algebra.

Our work extends prior work in learning hyperspherical representations and provides the following benefits:
\begin{enumerate}
    \item Our method naturally enforces the unitary constraint with the statistics during training, rather than relying on explicit Fourier coefficient normalization or clean-up
    memories when using deep learning representations for HRRs. Crucially, this is unsupervised, bridging the gap in the literature on embedding unstructured data into the vector symbolic framework.

    \item The independent von Mises distribution representation scales past dimensionalities where other hyperspherical methods struggle, avoiding posterior collapse that sampling from von Mises-Fisher and Power Spherical distributions at high $d$ can induce.

    \item The Clifford-VAE is competitive with Gaussian ($\mathcal{N}, \mathcal{N}_{L2}$) and Hyperspherical baselines ($\mathcal{S}$, $\mathcal{V}$) on evaluations. On CIFAR-10, our model improves over the strongest baseline ($\mathcal{N}$) for all evaluated dimensions on tests of  $d\in\{128, 256, \ldots, 4096\}$ and number of labeled data points, $n_{\ell}\in\{100, 600, 1000\}$, with an average increase in accuracy of $29\%$. Furthermore, on the simpler FashionMNIST dataset, we also observe best-in-class performance on supervised learning evaluations with limited data samples ($n_{\ell}=100$). 
    More crucially, across all experiments, our Clifford-VAE latent codes support effective binding and unbinding, as well as bundling with capacities matching or exceeding all existing methods for initializing VSAs. We show that our vectors perform better than canonical random HRR vectors on VSA tests, matching the state-of-the-art performance of FHRRs~\citep{Schlegel2020}. 
\end{enumerate}

\section{Results}
%%%%%%%%%%%%%%%%%%%%%%%%%%%%%%%%%%%%%%%%%%%%%%%%%%%%%%%

We evaluate the Clifford-VAE against a standard Gaussian VAE ($\mathcal{N}$), Gaussian VAE with L2
normalization ($\mathcal{N}_{L2}$), Power Spherical VAE ($\mathcal{S}$) from \citet{decao2020}, and von Mises-Fisher VAE
($\mathcal{V}$) on semi-supervised classification and standard VSA benchmarks, additionally observing that reconstruction quality remains highly competitive with Gaussian variants and superior to Hyperspherical baselines, especially when sampling at higher dimensions. These results expand upon the work of \citet{davidson2018,increasingexpressivitydavidson} by increasing latent representations into high-dimensional ($d>100$) representations, and incorporating convolutional encoding schemes.

\subsection{Mitigating Surface-Area Collapse of Hyperspherical Priors}

The \emph{hyperspherical bottleneck}~\citep{increasingexpressivitydavidson} arises from characterizing hyperspherical distributions with a scalar concentration parameter. In this setting, the surface area of the unit $(d{-}1)$-sphere $\mathcal{S}^{d-1}\subset\mathbb{R}^d$,
  \begin{equation}
  S(\mathcal{S}^{d-1}) = \frac{2\pi^{d/2}}{\Gamma(d/2)},
  \label{eq:sphere_area}
  \end{equation}
  first rises to a maximum at $d=7$ and thereafter decays to zero as $d\to\infty$, becoming
  vanishingly small beyond $d\approx17$~\citep{davidson2018, davidson2020shape}. When learning embeddings on a unit hypersphere with a single concentration parameter, the surface area of the hypersphere shrinks as the dimensionality increases, reducing the available representational space. 

\begin{figure}[H]
    \centering
    \includegraphics[width=0.7\linewidth]{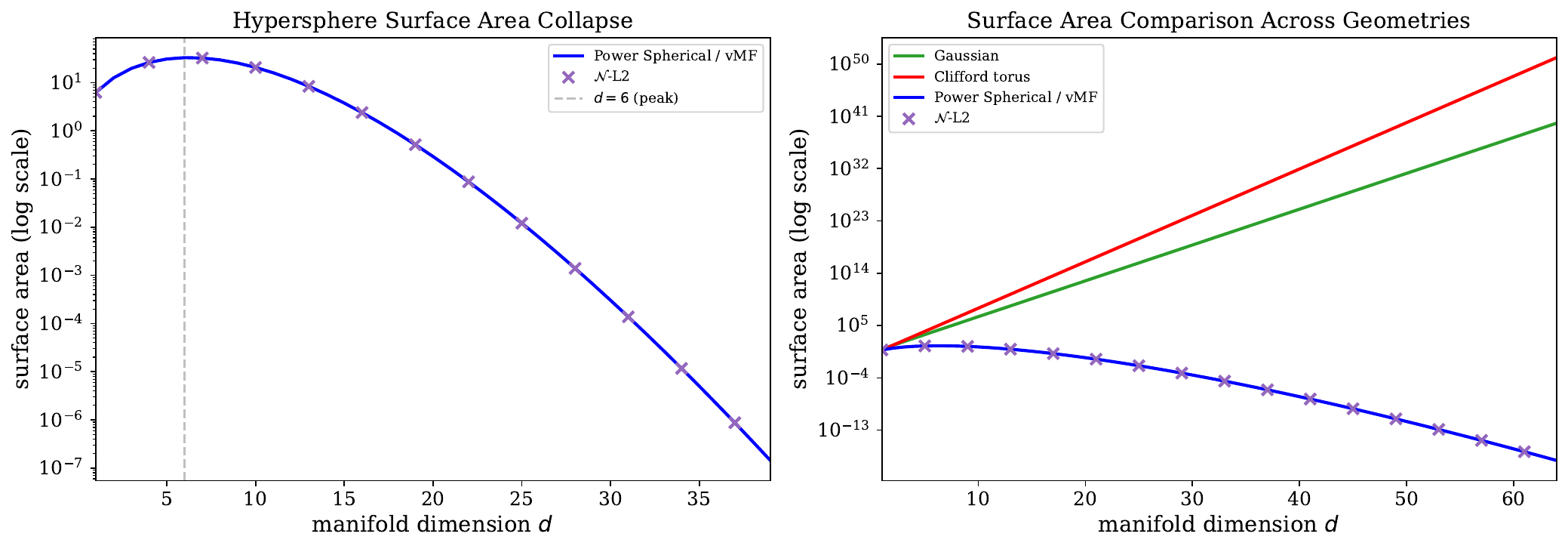}
\caption{\textit{Left:} surface area of the unit hypersphere $\mathcal{S}^{d-1}$ against
      dimension $d$. \textit{Right:} the same quantity compared across geometries --- the
      hypersphere (Power Spherical / vMF), the L2-normalized Gaussian, the unconstrained Gaussian,
      and the Clifford torus $(\mathcal{S}^1)^{d-1}$.}
      \label{fig:surface_area}
\end{figure}

Decomposing the hypersphere $\mathcal{S}^{d-1}$ into a product of lower-dimensional spheres
    $\mathcal{S}^{a_1}\times\cdots\times\mathcal{S}^{a_k}$ (with $\sum_i a_i = d-1$) alleviates
    this~\citep{increasingexpressivitydavidson}. The Clifford torus $(\mathcal{S}^1)^{d-1}$ is the maximal such
  factorization, as each factor has a measure independent of $d$. As a result, the total measure $(2\pi)^{d-1}$ grows with
  $d$ rather than vanishing (\Cref{fig:surface_area}). Thus, our method circumvents the \textit{hyperspherical bottleneck}, by learning an independent concentration $\kappa_i$ for each circle rather than relying on the limited expressivity of a single $\kappa$ that must be shared across all dimensions. This product space construction still implicitly enforces the unit-length constraint that enables effective bundling in the HRR, and of relevance to our motivation (learning high-dimensional representations for use in a VSA), allows for sampling at much higher values of $d$ than would be possible without our method's improved stability over hyperspherical baselines.

\subsection{Semi-supervised Learning (k-NN)}

\begin{figure}[H]
  \centering
  \begin{subfigure}[b]{0.48\linewidth}
    \includegraphics[width=\linewidth]{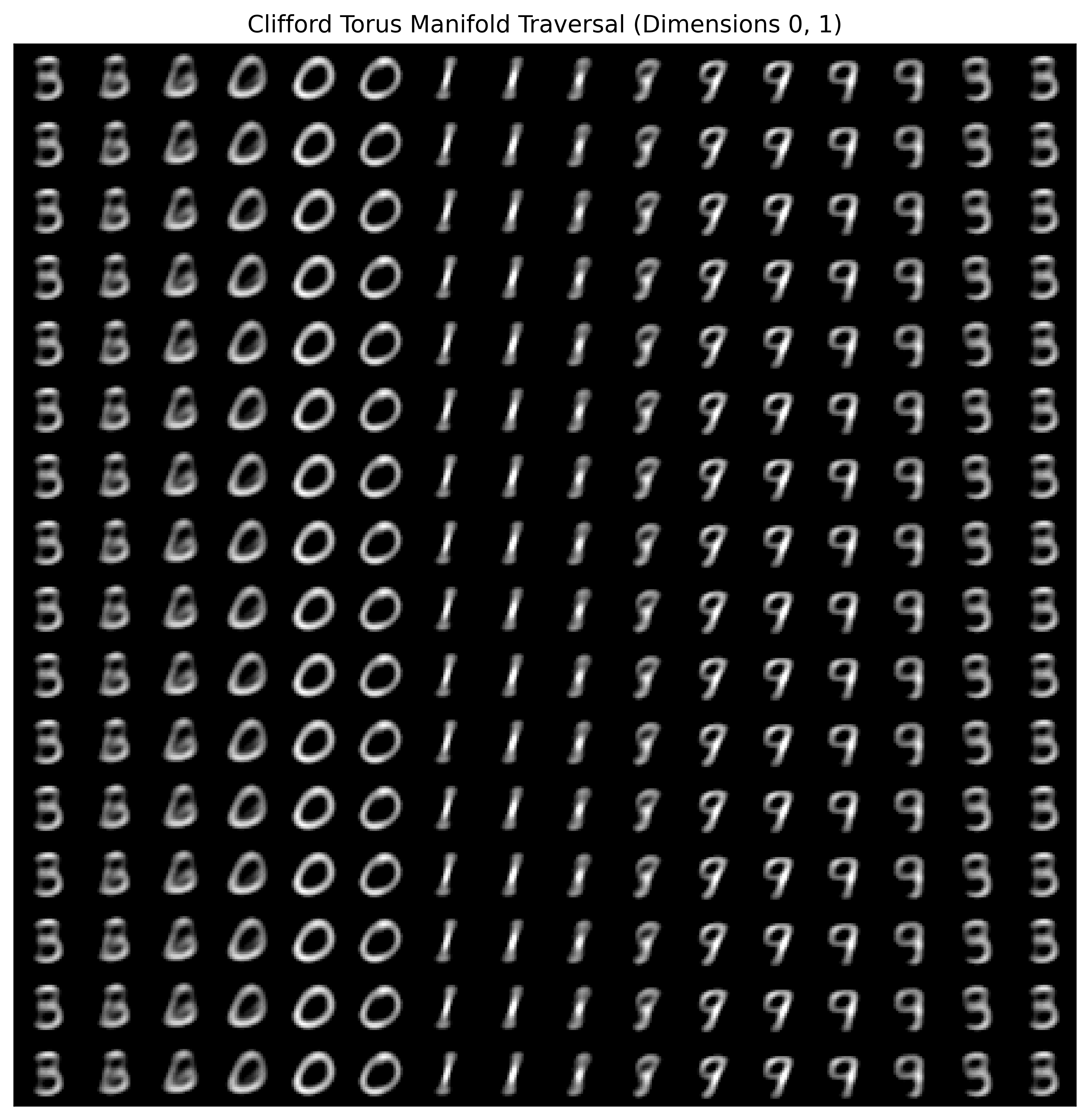}
    \caption{$d=2$.}
    \label{fig:mnist_manifold_d2}
  \end{subfigure}
  \hfill
  \begin{subfigure}[b]{0.48\linewidth}
    \includegraphics[width=\linewidth]{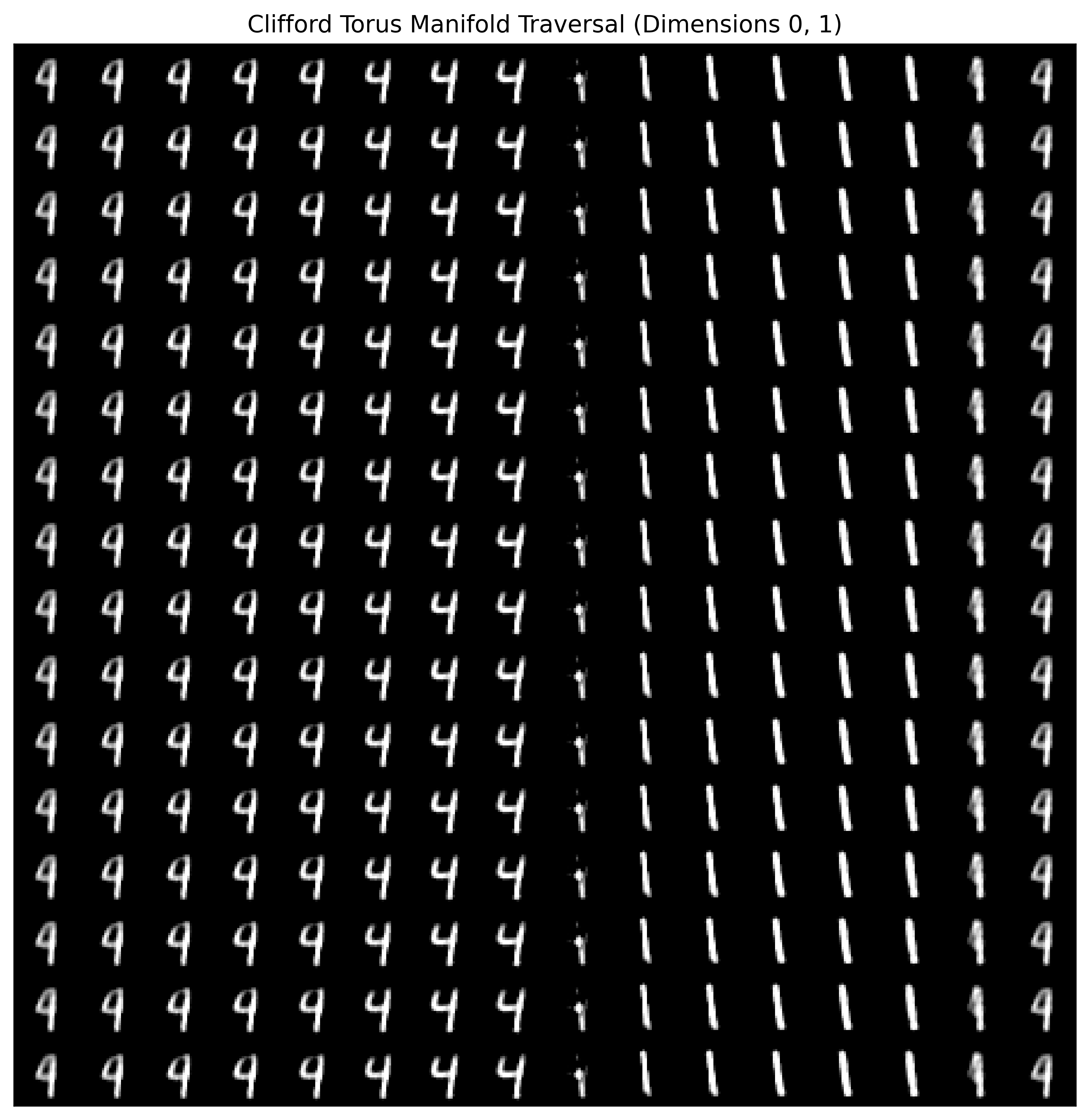}
    \caption{$d=5$ (traversing factors $0,1$).}
    \label{fig:mnist_manifold_d5}
  \end{subfigure}
  \caption{Clifford torus manifold traversals on MNIST. Each grid sweeps the
    first two latent angles $(\theta_0,\theta_1)$ over $[0,2\pi)^2$ and decodes samples from our Clifford distribution. At $d=5$ the remaining three factors are fixed for traversal.}
  \label{fig:mnist_manifold}
\end{figure}

Following \citet{davidson2018}, we train each VAE unsupervised and evaluate representations on a supervised learning task.
 A $k$-nearest-neighbor classifier with $k=5$ neighbors is fit on a small
  set of $n_{\ell}$ labeled training images, the `label budget'. \Cref{tab:mnist_knn} reports $k$-NN accuracy on MNIST across latent dimensions
  $d\in\{2,5,10,20,40,128\}$. The learned
  latent manifold is visualized in \Cref{fig:mnist_manifold}, where traversing the first two torus
  angles sweeps smoothly through digit classes across the periodic structure of $(\mathcal{S}^1)^{d-1}$.
  \Cref{tab:mnist_elbo} decomposes the test ELBO, showing the per-circle contribution as $d$ increases for
  our implementation, and the vanishing KL term for the Hyperspherical variants.

\begin{table}[H]
\centering
\caption{Semi-supervised $k$-NN accuracy (\%) on MNIST (20 runs), averaged over runs with the 95\% HDI of the mean (bootstrap, 20k resamples). Columns represent evaluations across $n_{\ell}\in\{100,600,1000\}$, the number of labeled training examples the $k$-NN classifier is fit on with $k=5$ neighbors. $\mathcal{N}$ = Gaussian, $\mathcal{N}_{L2}$ = L2-normalized Gaussian, $\mathcal{S}$ = Power Spherical, $\mathcal{V}$ = von Mises-Fisher, $\mathcal{C}$ = Clifford-VAE (ours). \textit{n.c.}\ = did not converge (vMF diverged at $d{=}128$). A clear (disjoint) winner is \textbf{\underline{bold and underlined}}; tied entries are \underline{underlined}.}
\label{tab:mnist_knn}
\resizebox{\textwidth}{!}{%
\begin{tabular}{l|ccccc|ccccc|ccccc}
\toprule
 & \multicolumn{5}{c|}{$n_{\ell}=100$} & \multicolumn{5}{c|}{$n_{\ell}=600$} & \multicolumn{5}{c}{$n_{\ell}=1000$} \\
 & $\mathcal{N}$ & $\mathcal{N}_{L2}$ & $\mathcal{S}$ & $\mathcal{V}$ & $\mathcal{C}$ (Ours) & $\mathcal{N}$ & $\mathcal{N}_{L2}$ & $\mathcal{S}$ & $\mathcal{V}$ & $\mathcal{C}$ (Ours) & $\mathcal{N}$ & $\mathcal{N}_{L2}$ & $\mathcal{S}$ & $\mathcal{V}$ & $\mathcal{C}$ (Ours) \\
\midrule
$d=2$ & \shortstack{\textbf{\underline{72.5}}\\{\tiny[70.8,\,74.3]}} & \shortstack{62.3\\{\tiny[60.3,\,64.3]}} & \shortstack{47.7\\{\tiny[47.2,\,48.3]}} & \shortstack{53.0\\{\tiny[51.8,\,54.3]}} & \shortstack{30.9\\{\tiny[29.2,\,32.4]}} & \shortstack{\textbf{\underline{83.4}}\\{\tiny[82.6,\,84.2]}} & \shortstack{67.7\\{\tiny[65.5,\,70.0]}} & \shortstack{50.9\\{\tiny[50.4,\,51.3]}} & \shortstack{57.7\\{\tiny[57.0,\,58.4]}} & \shortstack{31.5\\{\tiny[29.9,\,32.9]}} & \shortstack{\textbf{\underline{84.2}}\\{\tiny[83.6,\,84.9]}} & \shortstack{67.7\\{\tiny[65.6,\,69.9]}} & \shortstack{51.3\\{\tiny[50.9,\,51.7]}} & \shortstack{58.4\\{\tiny[57.8,\,59.0]}} & \shortstack{31.8\\{\tiny[30.4,\,33.2]}} \\
$d=5$ & \shortstack{\underline{84.2}\\{\tiny[83.2,\,85.2]}} & \shortstack{\underline{85.5}\\{\tiny[84.4,\,86.6]}} & \shortstack{65.6\\{\tiny[64.8,\,66.3]}} & \shortstack{70.9\\{\tiny[70.2,\,71.6]}} & \shortstack{66.1\\{\tiny[64.0,\,68.2]}} & \shortstack{\underline{92.5}\\{\tiny[92.3,\,92.6]}} & \shortstack{\underline{92.7}\\{\tiny[92.5,\,93.0]}} & \shortstack{76.1\\{\tiny[75.8,\,76.5]}} & \shortstack{81.3\\{\tiny[81.1,\,81.6]}} & \shortstack{79.3\\{\tiny[78.0,\,80.6]}} & \shortstack{\underline{93.1}\\{\tiny[93.0,\,93.3]}} & \shortstack{\underline{93.2}\\{\tiny[93.1,\,93.4]}} & \shortstack{77.9\\{\tiny[77.7,\,78.1]}} & \shortstack{82.5\\{\tiny[82.2,\,82.7]}} & \shortstack{81.0\\{\tiny[79.7,\,82.3]}} \\
$d=10$ & \shortstack{80.8\\{\tiny[79.7,\,82.0]}} & \shortstack{\textbf{\underline{85.0}}\\{\tiny[83.9,\,86.0]}} & \shortstack{64.0\\{\tiny[63.1,\,64.9]}} & \shortstack{71.7\\{\tiny[71.0,\,72.5]}} & \shortstack{72.1\\{\tiny[70.7,\,73.5]}} & \shortstack{92.0\\{\tiny[91.7,\,92.3]}} & \shortstack{\textbf{\underline{93.3}}\\{\tiny[93.1,\,93.5]}} & \shortstack{75.2\\{\tiny[74.9,\,75.4]}} & \shortstack{82.4\\{\tiny[82.0,\,82.8]}} & \shortstack{87.4\\{\tiny[86.9,\,87.9]}} & \shortstack{93.0\\{\tiny[92.8,\,93.1]}} & \shortstack{\textbf{\underline{94.0}}\\{\tiny[93.8,\,94.1]}} & \shortstack{77.1\\{\tiny[76.9,\,77.3]}} & \shortstack{84.2\\{\tiny[83.9,\,84.6]}} & \shortstack{89.3\\{\tiny[88.9,\,89.7]}} \\
$d=20$ & \shortstack{71.4\\{\tiny[69.8,\,72.9]}} & \shortstack{\textbf{\underline{75.8}}\\{\tiny[74.6,\,76.9]}} & \shortstack{58.3\\{\tiny[57.3,\,59.1]}} & \shortstack{67.0\\{\tiny[65.9,\,68.1]}} & \shortstack{68.1\\{\tiny[67.0,\,69.3]}} & \shortstack{89.7\\{\tiny[89.5,\,89.8]}} & \shortstack{\textbf{\underline{91.0}}\\{\tiny[90.8,\,91.2]}} & \shortstack{68.7\\{\tiny[67.9,\,69.4]}} & \shortstack{78.5\\{\tiny[78.0,\,79.0]}} & \shortstack{87.3\\{\tiny[86.8,\,87.7]}} & \shortstack{91.6\\{\tiny[91.4,\,91.9]}} & \shortstack{\textbf{\underline{92.6}}\\{\tiny[92.4,\,92.7]}} & \shortstack{70.6\\{\tiny[69.7,\,71.4]}} & \shortstack{79.9\\{\tiny[79.5,\,80.3]}} & \shortstack{89.3\\{\tiny[89.0,\,89.6]}} \\
$d=40$ & \shortstack{65.4\\{\tiny[63.7,\,66.9]}} & \shortstack{\underline{69.8}\\{\tiny[68.7,\,70.7]}} & \shortstack{49.9\\{\tiny[48.3,\,51.4]}} & \shortstack{57.2\\{\tiny[54.9,\,59.6]}} & \shortstack{\underline{68.6}\\{\tiny[67.3,\,70.0]}} & \shortstack{86.4\\{\tiny[86.0,\,86.7]}} & \shortstack{\textbf{\underline{89.1}}\\{\tiny[88.9,\,89.3]}} & \shortstack{55.7\\{\tiny[54.0,\,57.4]}} & \shortstack{67.9\\{\tiny[66.0,\,69.9]}} & \shortstack{87.8\\{\tiny[87.4,\,88.1]}} & \shortstack{89.1\\{\tiny[88.9,\,89.3]}} & \shortstack{\textbf{\underline{91.2}}\\{\tiny[91.0,\,91.3]}} & \shortstack{56.7\\{\tiny[55.0,\,58.5]}} & \shortstack{69.9\\{\tiny[68.2,\,71.6]}} & \shortstack{90.1\\{\tiny[89.8,\,90.4]}} \\
$d=128$ & \shortstack{50.6\\{\tiny[49.1,\,52.1]}} & \shortstack{\underline{67.2}\\{\tiny[66.2,\,68.2]}} & \shortstack{30.3\\{\tiny[29.3,\,31.2]}} & \textit{n.c.} & \shortstack{\underline{68.5}\\{\tiny[66.8,\,70.1]}} & \shortstack{73.8\\{\tiny[73.1,\,74.5]}} & \shortstack{\underline{86.5}\\{\tiny[86.2,\,86.8]}} & \shortstack{32.6\\{\tiny[32.2,\,33.1]}} & \textit{n.c.} & \shortstack{\underline{87.2}\\{\tiny[86.6,\,87.8]}} & \shortstack{78.5\\{\tiny[78.1,\,78.9]}} & \shortstack{\underline{89.1}\\{\tiny[88.9,\,89.3]}} & \shortstack{33.9\\{\tiny[33.2,\,34.6]}} & \textit{n.c.} & \shortstack{\underline{89.7}\\{\tiny[89.2,\,90.2]}} \\
\bottomrule
\end{tabular}%
}
\end{table}

\newpage
\begin{table}[H]
\centering
\caption{ELBO on MNIST (20 runs), averaged over runs with the 95\% HDI of the mean (bootstrap, 20k resamples). \textbf{Recon} $= -\log p_\theta(\mathbf{x}|\mathbf{z})$ is the reconstruction NLL (nats, $\downarrow$), \textbf{KL} is the divergence against the prior (nats), and $\mathbf{-ELBO}=\text{Recon}+\text{KL}$ is the total loss ($\downarrow$). In the $-$ELBO column a clear (disjoint) winner is \textbf{\underline{bold and underlined}}; tied leaders are \underline{underlined}. \textit{n.c.}\ = vMF did not converge at $d{=}128$. $\mathcal{N}$=Gaussian, $\mathcal{N}_{L2}$=L2-normalized Gaussian, $\mathcal{S}$=Power Spherical, $\mathcal{V}$=von Mises-Fisher, $\mathcal{C}$=Clifford-VAE (ours).}
\label{tab:mnist_elbo}
\resizebox{\textwidth}{!}{%
\begin{tabular}{l|ccc|ccc|ccc|ccc|ccc}
\toprule
 & \multicolumn{3}{c|}{$\mathcal{N}$} & \multicolumn{3}{c|}{$\mathcal{N}_{L2}$} & \multicolumn{3}{c|}{$\mathcal{S}$} & \multicolumn{3}{c|}{$\mathcal{V}$} & \multicolumn{3}{c}{$\mathcal{C}$ (Ours)} \\
 & Recon & KL & $-$ELBO & Recon & KL & $-$ELBO & Recon & KL & $-$ELBO & Recon & KL & $-$ELBO & Recon & KL & $-$ELBO \\
\midrule
$d=2$ & \shortstack{127.1\\{\tiny[126.8,\,127.4]}} & \shortstack{4.2\\{\tiny[4.1,\,4.2]}} & \shortstack{\textbf{\underline{131.3}}\\{\tiny[131.0,\,131.5]}} & \shortstack{148.8\\{\tiny[147.7,\,150.0]}} & \shortstack{3.9\\{\tiny[3.8,\,3.9]}} & \shortstack{152.7\\{\tiny[151.5,\,153.9]}} & \shortstack{177.6\\{\tiny[177.4,\,177.8]}} & \shortstack{1.5\\{\tiny[1.5,\,1.5]}} & \shortstack{179.1\\{\tiny[178.9,\,179.3]}} & \shortstack{168.9\\{\tiny[168.9,\,169.0]}} & \shortstack{2.0\\{\tiny[2.0,\,2.0]}} & \shortstack{170.9\\{\tiny[170.8,\,171.0]}} & \shortstack{184.0\\{\tiny[183.9,\,184.1]}} & \shortstack{1.2\\{\tiny[1.2,\,1.2]}} & \shortstack{185.2\\{\tiny[185.1,\,185.3]}} \\
$d=5$ & \shortstack{97.2\\{\tiny[97.0,\,97.4]}} & \shortstack{3.6\\{\tiny[3.6,\,3.6]}} & \shortstack{\textbf{\underline{100.8}}\\{\tiny[100.6,\,101.0]}} & \shortstack{104.0\\{\tiny[103.9,\,104.1]}} & \shortstack{3.9\\{\tiny[3.9,\,3.9]}} & \shortstack{107.9\\{\tiny[107.8,\,108.0]}} & \shortstack{175.4\\{\tiny[175.3,\,175.5]}} & \shortstack{1.7\\{\tiny[1.7,\,1.7]}} & \shortstack{177.1\\{\tiny[177.0,\,177.2]}} & \shortstack{161.8\\{\tiny[161.7,\,161.9]}} & \shortstack{2.5\\{\tiny[2.5,\,2.5]}} & \shortstack{164.3\\{\tiny[164.2,\,164.4]}} & \shortstack{135.9\\{\tiny[135.7,\,136.1]}} & \shortstack{5.0\\{\tiny[5.0,\,5.0]}} & \shortstack{140.9\\{\tiny[140.7,\,141.1]}} \\
$d=10$ & \shortstack{72.4\\{\tiny[72.2,\,72.6]}} & \shortstack{3.4\\{\tiny[3.4,\,3.4]}} & \shortstack{\textbf{\underline{75.8}}\\{\tiny[75.6,\,76.0]}} & \shortstack{74.8\\{\tiny[74.6,\,74.9]}} & \shortstack{4.2\\{\tiny[4.1,\,4.2]}} & \shortstack{78.9\\{\tiny[78.8,\,79.0]}} & \shortstack{179.4\\{\tiny[179.3,\,179.5]}} & \shortstack{1.6\\{\tiny[1.6,\,1.6]}} & \shortstack{181.0\\{\tiny[180.9,\,181.1]}} & \shortstack{166.6\\{\tiny[166.5,\,166.8]}} & \shortstack{2.4\\{\tiny[2.4,\,2.4]}} & \shortstack{169.1\\{\tiny[168.9,\,169.2]}} & \shortstack{105.5\\{\tiny[105.3,\,105.6]}} & \shortstack{11.2\\{\tiny[11.2,\,11.2]}} & \shortstack{116.7\\{\tiny[116.5,\,116.9]}} \\
$d=20$ & \shortstack{54.7\\{\tiny[54.6,\,54.8]}} & \shortstack{3.2\\{\tiny[3.2,\,3.2]}} & \shortstack{\textbf{\underline{57.9}}\\{\tiny[57.7,\,58.0]}} & \shortstack{55.8\\{\tiny[55.7,\,55.9]}} & \shortstack{4.3\\{\tiny[4.3,\,4.3]}} & \shortstack{60.1\\{\tiny[60.1,\,60.2]}} & \shortstack{185.1\\{\tiny[185.0,\,185.2]}} & \shortstack{1.2\\{\tiny[1.2,\,1.2]}} & \shortstack{186.4\\{\tiny[186.3,\,186.5]}} & \shortstack{177.0\\{\tiny[176.9,\,177.2]}} & \shortstack{1.9\\{\tiny[1.9,\,1.9]}} & \shortstack{178.9\\{\tiny[178.8,\,179.0]}} & \shortstack{81.2\\{\tiny[81.0,\,81.3]}} & \shortstack{23.7\\{\tiny[23.7,\,23.7]}} & \shortstack{104.9\\{\tiny[104.7,\,105.0]}} \\
$d=40$ & \shortstack{38.6\\{\tiny[38.5,\,38.7]}} & \shortstack{3.2\\{\tiny[3.1,\,3.2]}} & \shortstack{\textbf{\underline{41.7}}\\{\tiny[41.6,\,41.8]}} & \shortstack{39.2\\{\tiny[39.1,\,39.3]}} & \shortstack{4.6\\{\tiny[4.6,\,4.6]}} & \shortstack{43.8\\{\tiny[43.7,\,43.9]}} & \shortstack{191.3\\{\tiny[191.2,\,191.4]}} & \shortstack{0.8\\{\tiny[0.8,\,0.8]}} & \shortstack{192.1\\{\tiny[192.0,\,192.2]}} & \shortstack{190.2\\{\tiny[187.4,\,193.0]}} & \shortstack{-1.1\\{\tiny[-3.2,\,1.1]}} & \shortstack{189.1\\{\tiny[188.6,\,189.7]}} & \shortstack{63.1\\{\tiny[63.0,\,63.3]}} & \shortstack{48.7\\{\tiny[48.7,\,48.7]}} & \shortstack{111.8\\{\tiny[111.7,\,112.0]}} \\
$d=128$ & \shortstack{16.0\\{\tiny[15.8,\,16.3]}} & \shortstack{2.5\\{\tiny[2.5,\,2.5]}} & \shortstack{\textbf{\underline{18.5}}\\{\tiny[18.3,\,18.8]}} & \shortstack{25.4\\{\tiny[25.0,\,25.8]}} & \shortstack{3.4\\{\tiny[3.4,\,3.5]}} & \shortstack{28.8\\{\tiny[28.5,\,29.2]}} & \shortstack{199.8\\{\tiny[199.7,\,199.8]}} & \shortstack{0.3\\{\tiny[0.3,\,0.3]}} & \shortstack{200.1\\{\tiny[200.0,\,200.2]}} & \textit{n.c.} & \textit{n.c.} & \textit{n.c.} & \shortstack{50.6\\{\tiny[42.4,\,58.9]}} & \shortstack{139.0\\{\tiny[119.5,\,158.6]}} & \shortstack{189.6\\{\tiny[178.2,\,201.0]}} \\
\bottomrule
\end{tabular}%
}
\end{table}

\Cref{tab:fashion_knn_full_hdi} reports $k$-NN accuracy on FashionMNIST, where the Clifford-VAE
  is best-in-class at the smallest label budget ($n_\ell=100$). On the CIFAR-10 dataset, \Cref{tab:cifar_knn_hdi}
  shows that Clifford-VAE improves over the strongest baseline ($\mathcal{N}$) at every dimension and
  label budget evaluated.
% fashionMNIST
\begin{table}[H]
\centering
\caption{Semi-supervised $k$-NN accuracy (\%) on FashionMNIST (30 runs). $\mathcal{N}$ = Gaussian, $\mathcal{N}_{L2}$ = L2-normalized Gaussian, $\mathcal{S}$ = Power Spherical, $\mathcal{V}$ = von Mises-Fisher, $\mathcal{C}$ = Clifford-VAE (ours). }
\label{tab:fashion_knn_full_hdi}
\resizebox{\textwidth}{!}{%
\begin{tabular}{l|cccc|cccc|cccc}
\toprule
 & \multicolumn{4}{c|}{$n_{\ell}=100$} & \multicolumn{4}{c|}{$n_{\ell}=600$} & \multicolumn{4}{c}{$n_{\ell}=1000$} \\
 & $\mathcal{N}$ & $\mathcal{N}_{L2}$ & $\mathcal{S}$ & $\mathcal{C}$ (Ours) & $\mathcal{N}$ & $\mathcal{N}_{L2}$ & $\mathcal{S}$ & $\mathcal{C}$ (Ours) & $\mathcal{N}$ & $\mathcal{N}_{L2}$ & $\mathcal{S}$ & $\mathcal{C}$ (Ours) \\
\midrule
$d=128$ & \shortstack{50.2\\{\tiny[48.6,\,51.6]}} & \shortstack{58.8\\{\tiny[58.1,\,59.5]}} & \shortstack{46.1\\{\tiny[45.3,\,46.9]}} & \shortstack{\textbf{\underline{63.5}}\\{\tiny[62.7,\,64.3]}} & \shortstack{70.3\\{\tiny[69.8,\,70.8]}} & \shortstack{75.2\\{\tiny[74.8,\,75.5]}} & \shortstack{56.1\\{\tiny[55.8,\,56.4]}} & \shortstack{\textbf{\underline{76.4}}\\{\tiny[76.2,\,76.6]}} & \shortstack{74.5\\{\tiny[74.2,\,74.9]}} & \shortstack{77.6\\{\tiny[77.3,\,77.8]}} & \shortstack{58.0\\{\tiny[57.5,\,58.4]}} & \shortstack{\textbf{\underline{78.4}}\\{\tiny[78.2,\,78.6]}} \\
$d=256$ & \shortstack{50.4\\{\tiny[49.4,\,51.5]}} & \shortstack{59.4\\{\tiny[58.3,\,60.4]}} & \shortstack{42.1\\{\tiny[41.3,\,42.9]}} & \shortstack{\textbf{\underline{63.9}}\\{\tiny[63.1,\,64.7]}} & \shortstack{70.8\\{\tiny[70.3,\,71.2]}} & \shortstack{\underline{76.3}\\{\tiny[76.0,\,76.6]}} & \shortstack{51.7\\{\tiny[51.2,\,52.2]}} & \shortstack{\underline{76.1}\\{\tiny[75.9,\,76.3]}} & \shortstack{74.8\\{\tiny[74.4,\,75.2]}} & \shortstack{\textbf{\underline{78.7}}\\{\tiny[78.5,\,78.8]}} & \shortstack{55.6\\{\tiny[55.1,\,56.0]}} & \shortstack{78.1\\{\tiny[77.9,\,78.3]}} \\
$d=512$ & \shortstack{50.3\\{\tiny[49.3,\,51.2]}} & \shortstack{\underline{61.3}\\{\tiny[60.5,\,62.2]}} & \shortstack{42.4\\{\tiny[41.8,\,43.1]}} & \shortstack{\underline{62.9}\\{\tiny[61.9,\,63.9]}} & \shortstack{71.1\\{\tiny[70.7,\,71.6]}} & \shortstack{\textbf{\underline{76.9}}\\{\tiny[76.7,\,77.2]}} & \shortstack{51.7\\{\tiny[51.0,\,52.4]}} & \shortstack{75.8\\{\tiny[75.5,\,76.0]}} & \shortstack{74.8\\{\tiny[74.4,\,75.1]}} & \shortstack{\textbf{\underline{79.2}}\\{\tiny[79.0,\,79.4]}} & \shortstack{55.7\\{\tiny[55.1,\,56.2]}} & \shortstack{77.5\\{\tiny[77.3,\,77.7]}} \\
$d=1024$ & \shortstack{50.9\\{\tiny[49.3,\,52.5]}} & \shortstack{\underline{60.9}\\{\tiny[59.9,\,61.9]}} & \shortstack{41.4\\{\tiny[40.6,\,42.2]}} & \shortstack{\underline{62.9}\\{\tiny[61.9,\,63.9]}} & \shortstack{71.8\\{\tiny[71.2,\,72.5]}} & \shortstack{\textbf{\underline{76.4}}\\{\tiny[76.2,\,76.7]}} & \shortstack{52.4\\{\tiny[51.8,\,52.9]}} & \shortstack{75.1\\{\tiny[74.8,\,75.4]}} & \shortstack{75.5\\{\tiny[75.1,\,75.8]}} & \shortstack{\textbf{\underline{78.5}}\\{\tiny[78.3,\,78.7]}} & \shortstack{56.4\\{\tiny[55.9,\,56.9]}} & \shortstack{77.0\\{\tiny[76.8,\,77.2]}} \\
$d=2048$ & \shortstack{51.6\\{\tiny[50.3,\,52.9]}} & \shortstack{\underline{60.7}\\{\tiny[59.8,\,61.6]}} & \shortstack{44.7\\{\tiny[43.8,\,45.5]}} & \shortstack{\underline{61.4}\\{\tiny[60.6,\,62.2]}} & \shortstack{72.7\\{\tiny[72.2,\,73.2]}} & \shortstack{\textbf{\underline{75.1}}\\{\tiny[74.8,\,75.3]}} & \shortstack{57.8\\{\tiny[57.3,\,58.2]}} & \shortstack{74.2\\{\tiny[74.0,\,74.5]}} & \shortstack{76.3\\{\tiny[75.8,\,76.7]}} & \shortstack{\textbf{\underline{77.6}}\\{\tiny[77.3,\,77.9]}} & \shortstack{60.2\\{\tiny[59.8,\,60.7]}} & \shortstack{76.2\\{\tiny[76.0,\,76.4]}} \\
$d=4096$ & \shortstack{54.9\\{\tiny[53.8,\,55.9]}} & \shortstack{\underline{59.8}\\{\tiny[59.1,\,60.6]}} & \shortstack{46.2\\{\tiny[45.1,\,47.3]}} & \shortstack{\underline{61.0}\\{\tiny[59.8,\,62.2]}} & \shortstack{66.8\\{\tiny[65.8,\,67.9]}} & \shortstack{\underline{74.5}\\{\tiny[74.3,\,74.8]}} & \shortstack{59.1\\{\tiny[58.6,\,59.5]}} & \shortstack{\underline{74.5}\\{\tiny[74.1,\,74.9]}} & \shortstack{69.0\\{\tiny[68.0,\,70.2]}} & \shortstack{\underline{76.7}\\{\tiny[76.5,\,77.0]}} & \shortstack{61.3\\{\tiny[60.9,\,61.8]}} & \shortstack{\underline{76.2}\\{\tiny[75.7,\,76.8]}} \\
\bottomrule
\end{tabular}%
}
\end{table}

\begin{table}[H]
\centering
\caption{Semi-supervised $k$-NN accuracy (\%) on CIFAR-10 (30 runs), averaged over runs with the 95\% HDI of the mean (bootstrap, 20k resamples). $\mathcal{N}$ = Gaussian, $\mathcal{N}_{L2}$ = L2-normalized Gaussian, $\mathcal{S}$ = Power Spherical, $\mathcal{C}$ = Clifford-VAE (ours). A clear (disjoint) winner is \textbf{\underline{bold and underlined}}; tied leaders are \underline{underlined}.}
\label{tab:cifar_knn_hdi}
\resizebox{\textwidth}{!}{%
\begin{tabular}{l|cccc|cccc|cccc}
\toprule
 & \multicolumn{4}{c|}{$n_{\ell}=100$} & \multicolumn{4}{c|}{$n_{\ell}=600$} & \multicolumn{4}{c}{$n_{\ell}=1000$} \\
 & $\mathcal{N}$ & $\mathcal{N}_{L2}$ & $\mathcal{S}$ & $\mathcal{C}$ (Ours) & $\mathcal{N}$ & $\mathcal{N}_{L2}$ & $\mathcal{S}$ & $\mathcal{C}$ (Ours) & $\mathcal{N}$ & $\mathcal{N}_{L2}$ & $\mathcal{S}$ & $\mathcal{C}$ (Ours) \\
\midrule
$d=128$ & \shortstack{14.5\\{\tiny[13.3,\,15.8]}} & \shortstack{9.7\\{\tiny[8.5,\,10.9]}} & \shortstack{10.2\\{\tiny[9.0,\,11.5]}} & \shortstack{\textbf{\underline{21.6}}\\{\tiny[19.8,\,23.4]}} & \shortstack{20.6\\{\tiny[20.0,\,21.1]}} & \shortstack{9.6\\{\tiny[9.2,\,9.9]}} & \shortstack{9.8\\{\tiny[9.4,\,10.2]}} & \shortstack{\textbf{\underline{27.2}}\\{\tiny[26.6,\,27.8]}} & \shortstack{22.4\\{\tiny[22.0,\,22.8]}} & \shortstack{10.1\\{\tiny[9.8,\,10.4]}} & \shortstack{10.1\\{\tiny[9.8,\,10.4]}} & \shortstack{\textbf{\underline{29.9}}\\{\tiny[29.3,\,30.4]}} \\
$d=256$ & \shortstack{14.4\\{\tiny[12.8,\,16.0]}} & \shortstack{9.2\\{\tiny[8.2,\,10.1]}} & \shortstack{9.4\\{\tiny[8.3,\,10.6]}} & \shortstack{\textbf{\underline{18.4}}\\{\tiny[17.0,\,19.6]}} & \shortstack{17.1\\{\tiny[16.3,\,17.9]}} & \shortstack{9.8\\{\tiny[9.5,\,10.1]}} & \shortstack{9.8\\{\tiny[9.4,\,10.2]}} & \shortstack{\textbf{\underline{24.6}}\\{\tiny[24.0,\,25.1]}} & \shortstack{18.4\\{\tiny[17.9,\,18.9]}} & \shortstack{9.8\\{\tiny[9.5,\,10.0]}} & \shortstack{10.1\\{\tiny[9.8,\,10.3]}} & \shortstack{\textbf{\underline{26.7}}\\{\tiny[26.2,\,27.3]}} \\
$d=512$ & \shortstack{12.8\\{\tiny[11.7,\,13.9]}} & \shortstack{9.1\\{\tiny[8.2,\,10.0]}} & \shortstack{8.9\\{\tiny[8.0,\,9.8]}} & \shortstack{\textbf{\underline{16.8}}\\{\tiny[15.2,\,18.4]}} & \shortstack{14.7\\{\tiny[14.3,\,15.1]}} & \shortstack{10.1\\{\tiny[9.6,\,10.6]}} & \shortstack{9.9\\{\tiny[9.5,\,10.2]}} & \shortstack{\textbf{\underline{19.7}}\\{\tiny[19.0,\,20.5]}} & \shortstack{15.5\\{\tiny[15.1,\,15.9]}} & \shortstack{9.8\\{\tiny[9.4,\,10.2]}} & \shortstack{10.1\\{\tiny[9.8,\,10.5]}} & \shortstack{\textbf{\underline{21.3}}\\{\tiny[20.9,\,21.7]}} \\
$d=1024$ & \shortstack{11.3\\{\tiny[10.3,\,12.3]}} & \shortstack{8.2\\{\tiny[7.4,\,9.0]}} & \shortstack{9.3\\{\tiny[8.3,\,10.4]}} & \shortstack{\textbf{\underline{14.4}}\\{\tiny[13.3,\,15.4]}} & \shortstack{13.0\\{\tiny[12.5,\,13.4]}} & \shortstack{10.0\\{\tiny[9.6,\,10.4]}} & \shortstack{9.5\\{\tiny[9.0,\,9.9]}} & \shortstack{\textbf{\underline{16.6}}\\{\tiny[16.2,\,17.0]}} & \shortstack{13.5\\{\tiny[13.1,\,13.9]}} & \shortstack{10.1\\{\tiny[9.8,\,10.4]}} & \shortstack{9.8\\{\tiny[9.5,\,10.0]}} & \shortstack{\textbf{\underline{17.6}}\\{\tiny[17.2,\,18.0]}} \\
$d=2048$ & \shortstack{11.1\\{\tiny[10.0,\,12.2]}} & \shortstack{10.0\\{\tiny[9.0,\,11.0]}} & \shortstack{8.9\\{\tiny[7.8,\,9.9]}} & \shortstack{\textbf{\underline{13.9}}\\{\tiny[12.5,\,15.3]}} & \shortstack{12.5\\{\tiny[12.1,\,13.0]}} & \shortstack{9.7\\{\tiny[9.3,\,10.1]}} & \shortstack{9.8\\{\tiny[9.3,\,10.2]}} & \shortstack{\textbf{\underline{14.3}}\\{\tiny[13.8,\,14.9]}} & \shortstack{12.7\\{\tiny[12.3,\,13.0]}} & \shortstack{9.8\\{\tiny[9.4,\,10.1]}} & \shortstack{9.4\\{\tiny[9.1,\,9.8]}} & \shortstack{\textbf{\underline{14.9}}\\{\tiny[14.5,\,15.4]}} \\
$d=4096$ & \shortstack{\underline{10.3}\\{\tiny[9.3,\,11.3]}} & \shortstack{\underline{10.1}\\{\tiny[9.4,\,10.9]}} & \shortstack{9.4\\{\tiny[8.3,\,10.5]}} & \shortstack{\underline{12.2}\\{\tiny[10.9,\,13.4]}} & \shortstack{11.5\\{\tiny[11.1,\,11.9]}} & \shortstack{10.1\\{\tiny[9.6,\,10.5]}} & \shortstack{9.9\\{\tiny[9.5,\,10.4]}} & \shortstack{\textbf{\underline{13.1}}\\{\tiny[12.5,\,13.6]}} & \shortstack{11.9\\{\tiny[11.5,\,12.2]}} & \shortstack{9.9\\{\tiny[9.6,\,10.2]}} & \shortstack{9.7\\{\tiny[9.4,\,10.0]}} & \shortstack{\textbf{\underline{13.7}}\\{\tiny[13.4,\,14.0]}} \\
\bottomrule
\end{tabular}%
}
\end{table}

% ===== cifar-10 fid
\begin{table}[H]
\centering
\caption{Generation FID on CIFAR-10 (30 runs), averaged over runs with the 95\% HDI of the mean (bootstrap, 20k resamples); lower is better ($\downarrow$). $\mathcal{N}$ = Gaussian, $\mathcal{N}_{L2}$ = L2-normalized Gaussian, $\mathcal{S}$ = Power Spherical, $\mathcal{C}$ = Clifford-VAE.}
\label{tab:cifar_fid_hdi}
\begin{tabular}{lcccc}
\toprule
 & $\mathcal{N}$ & $\mathcal{N}_{L2}$ & $\mathcal{S}$ & $\mathcal{C}$ (Ours) \\
\midrule
$d=128$ & \shortstack{\textbf{\underline{136.6}}\\{\tiny[136.1,\,137.2]}} & \shortstack{159.1\\{\tiny[158.3,\,159.9]}} & \shortstack{411.7\\{\tiny[408.5,\,415.0]}} & \shortstack{140.9\\{\tiny[140.5,\,141.4]}} \\
$d=256$ & \shortstack{\textbf{\underline{135.1}}\\{\tiny[134.7,\,135.6]}} & \shortstack{158.0\\{\tiny[156.9,\,159.1]}} & \shortstack{422.3\\{\tiny[418.5,\,425.9]}} & \shortstack{149.2\\{\tiny[148.6,\,150.0]}} \\
$d=512$ & \shortstack{\textbf{\underline{135.5}}\\{\tiny[135.0,\,136.0]}} & \shortstack{173.5\\{\tiny[172.6,\,174.5]}} & \shortstack{418.4\\{\tiny[412.5,\,424.2]}} & \shortstack{193.7\\{\tiny[192.5,\,195.0]}} \\
$d=1024$ & \shortstack{\textbf{\underline{138.2}}\\{\tiny[137.6,\,138.8]}} & \shortstack{178.0\\{\tiny[177.2,\,178.7]}} & \shortstack{422.0\\{\tiny[415.6,\,428.0]}} & \shortstack{216.9\\{\tiny[215.5,\,218.2]}} \\
$d=2048$ & \shortstack{\textbf{\underline{140.4}}\\{\tiny[139.8,\,140.9]}} & \shortstack{153.5\\{\tiny[152.2,\,154.7]}} & \shortstack{423.5\\{\tiny[420.3,\,426.8]}} & \shortstack{226.7\\{\tiny[224.7,\,228.7]}} \\
$d=4096$ & \shortstack{\textbf{\underline{140.9}}\\{\tiny[140.2,\,141.5]}} & \shortstack{149.7\\{\tiny[148.5,\,151.0]}} & \shortstack{430.3\\{\tiny[427.1,\,433.6]}} & \shortstack{222.0\\{\tiny[220.1,\,223.9]}} \\
\bottomrule
\end{tabular}
\end{table}

\begin{figure}[H]
    \centering
    \begin{subfigure}[b]{0.45\linewidth}
      \includegraphics[width=\linewidth]{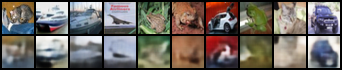}
      \caption{Clifford (Ours, $d=256$).}
      \label{fig:cifar_recon_clifford}
    \end{subfigure}   
    \hfill
    \begin{subfigure}[b]{0.45\linewidth}
      \includegraphics[width=\linewidth]{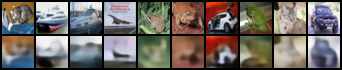}
      \caption{Gaussian, L2-normalized ($d=256$).}
      \label{fig:cifar_recon_gaussian}
    \end{subfigure}   

    \vspace{2pt}

    \begin{subfigure}[b]{0.45\linewidth}
      \includegraphics[width=\linewidth]{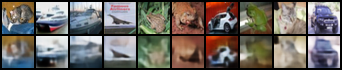}
      \caption{Gaussian, no L2 normalization ($d=256$).}
      \label{fig:cifar_recon_gaussian_nol2}
    \end{subfigure}
    \hfill
    \begin{subfigure}[b]{0.45\linewidth}
      \includegraphics[width=\linewidth]{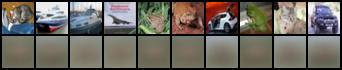}
      \caption{Power Spherical ($d=256$).}
      \label{fig:cifar_recon_powerspherical}
    \end{subfigure}
    \caption{CIFAR-10 reconstructions at $d=256$. Clifford reconstructions are qualitatively
      comparable to all baselines despite the unit-magnitude constraint on Fourier coefficients.}
    \label{fig:cifar_reconstructions}
  \end{figure}

\Cref{tab:cifar_fid_hdi} reports generation FID on CIFAR-10. The Gaussian VAE attains the lowest
FID; the Clifford-VAE trades some reconstruction fidelity for its unit-magnitude constraint but
degrades far more gracefully with dimension than the Power Spherical baseline. Qualitatively,
\Cref{fig:cifar_reconstructions} shows Clifford reconstructions remain comparable to all Gaussian baselines evaluated
at $d=256$.

\begin{figure}[H]
    \centering
    \begin{subfigure}[b]{0.45\linewidth}
      \includegraphics[width=\linewidth]{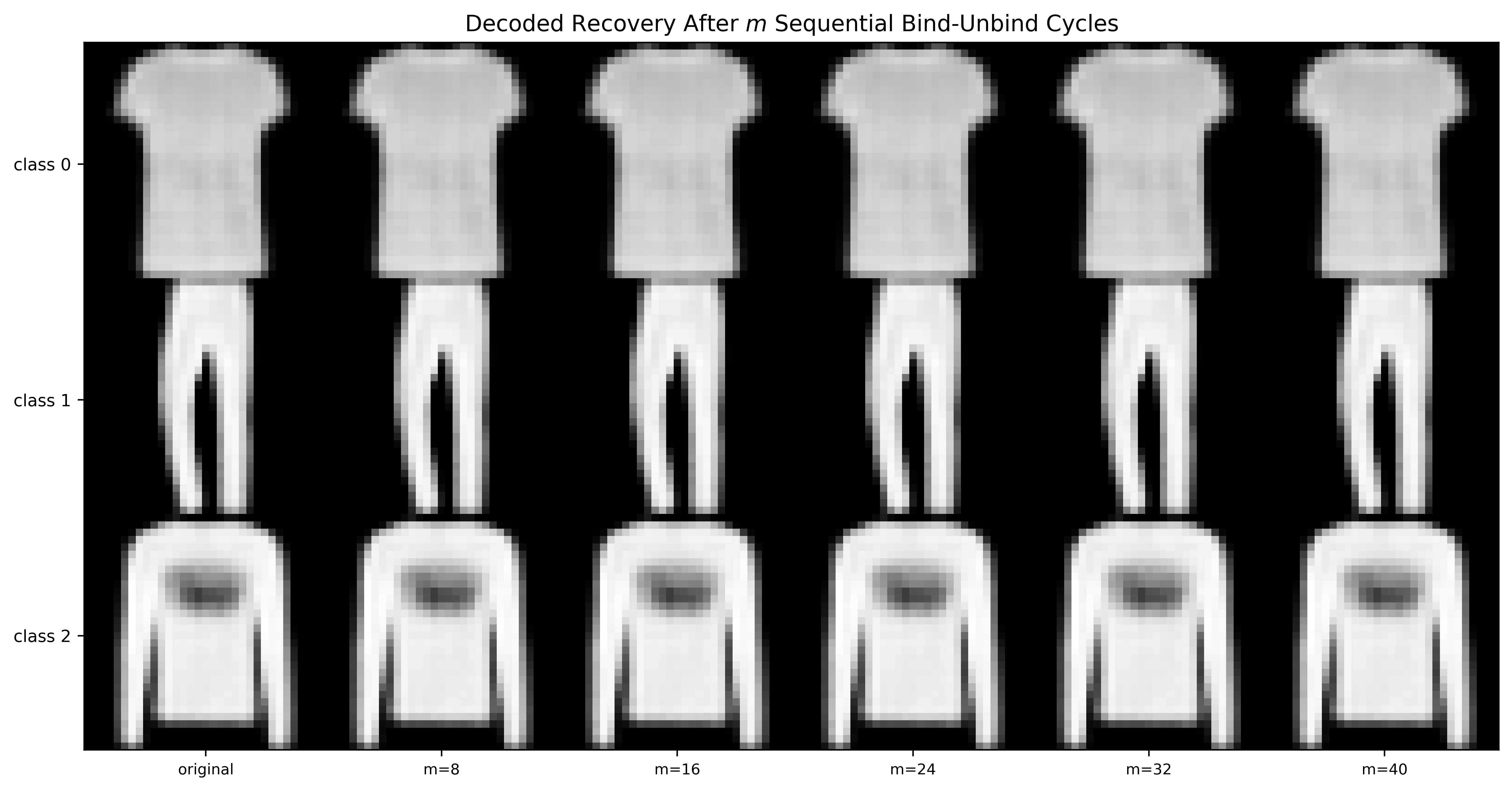}
      \caption{Clifford ($*$).}
      \label{fig:fashion_rec_cliff_inv}
    \end{subfigure}
    \hfill
    \begin{subfigure}[b]{0.45\linewidth}
      \includegraphics[width=\linewidth]{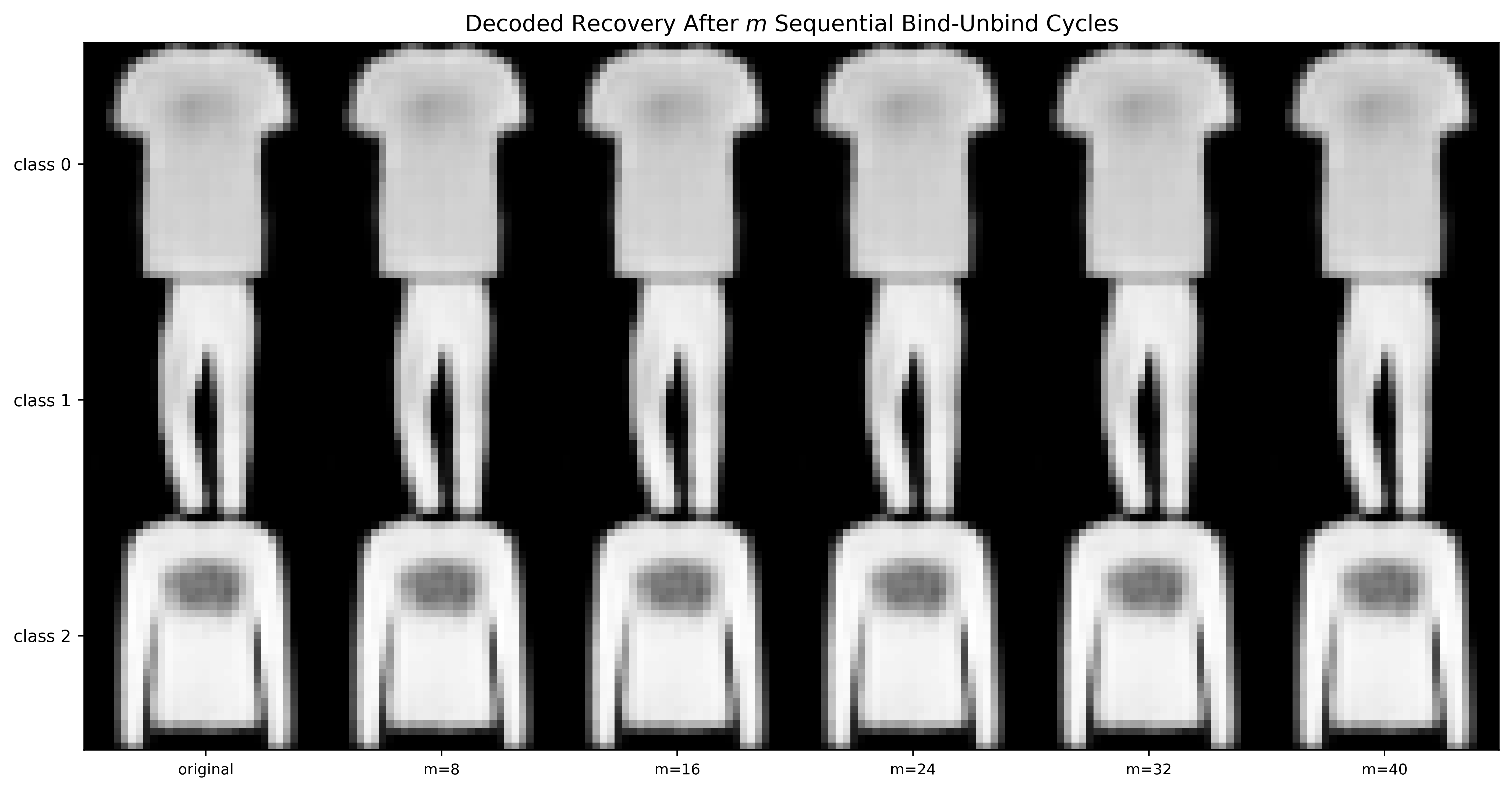}
      \caption{Clifford ($\dagger$).}
      \label{fig:fashion_rec_cliff_dag}
    \end{subfigure}
    \vspace{4pt}
    \begin{subfigure}[b]{0.45\linewidth}
      \includegraphics[width=\linewidth]{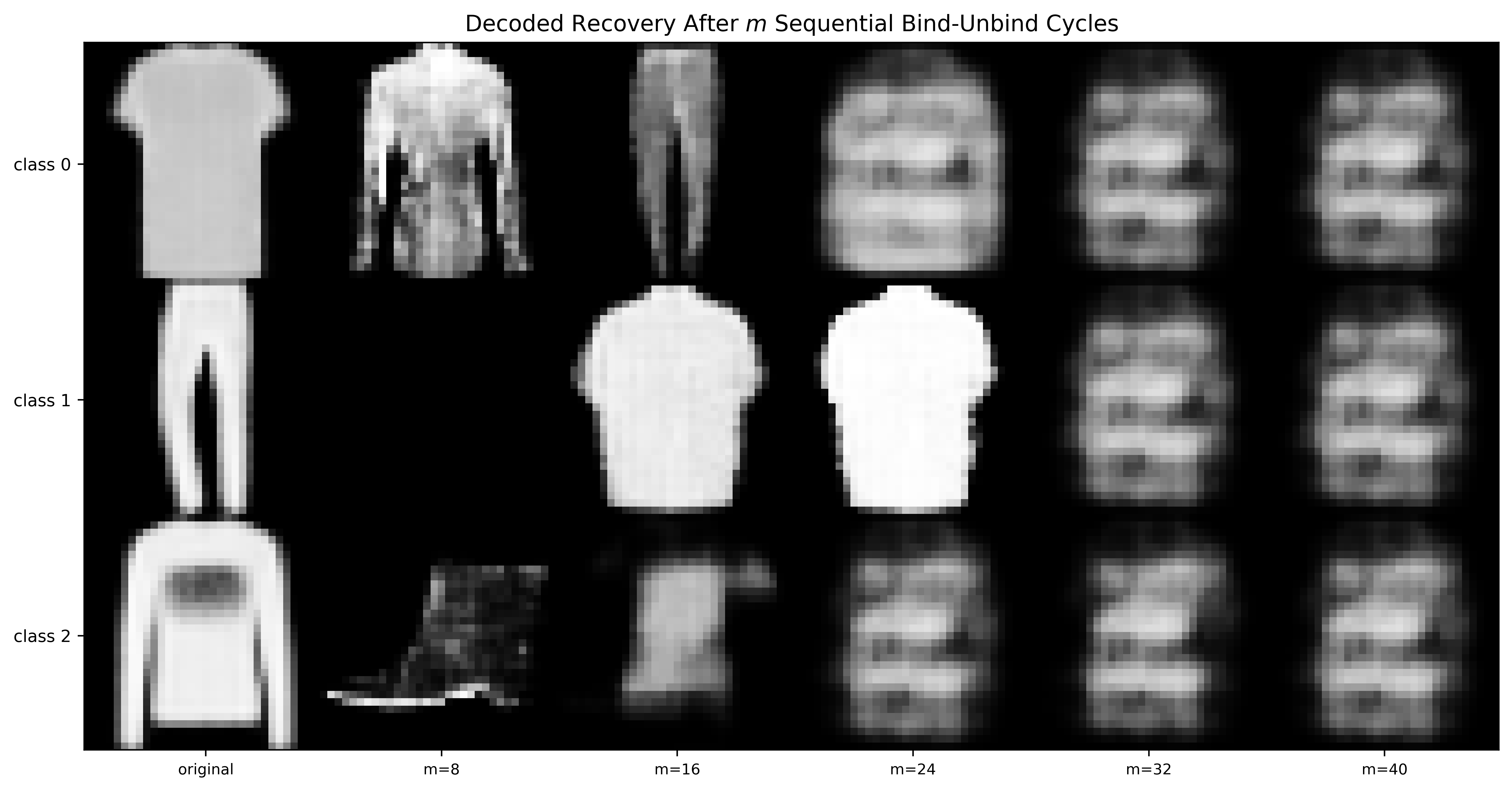}
      \caption{Gaussian L2 ($*$).}
      \label{fig:fashion_rec_gauss_inv} 
    \end{subfigure}
    \hfill
    \begin{subfigure}[b]{0.45\linewidth}
      \includegraphics[width=\linewidth]{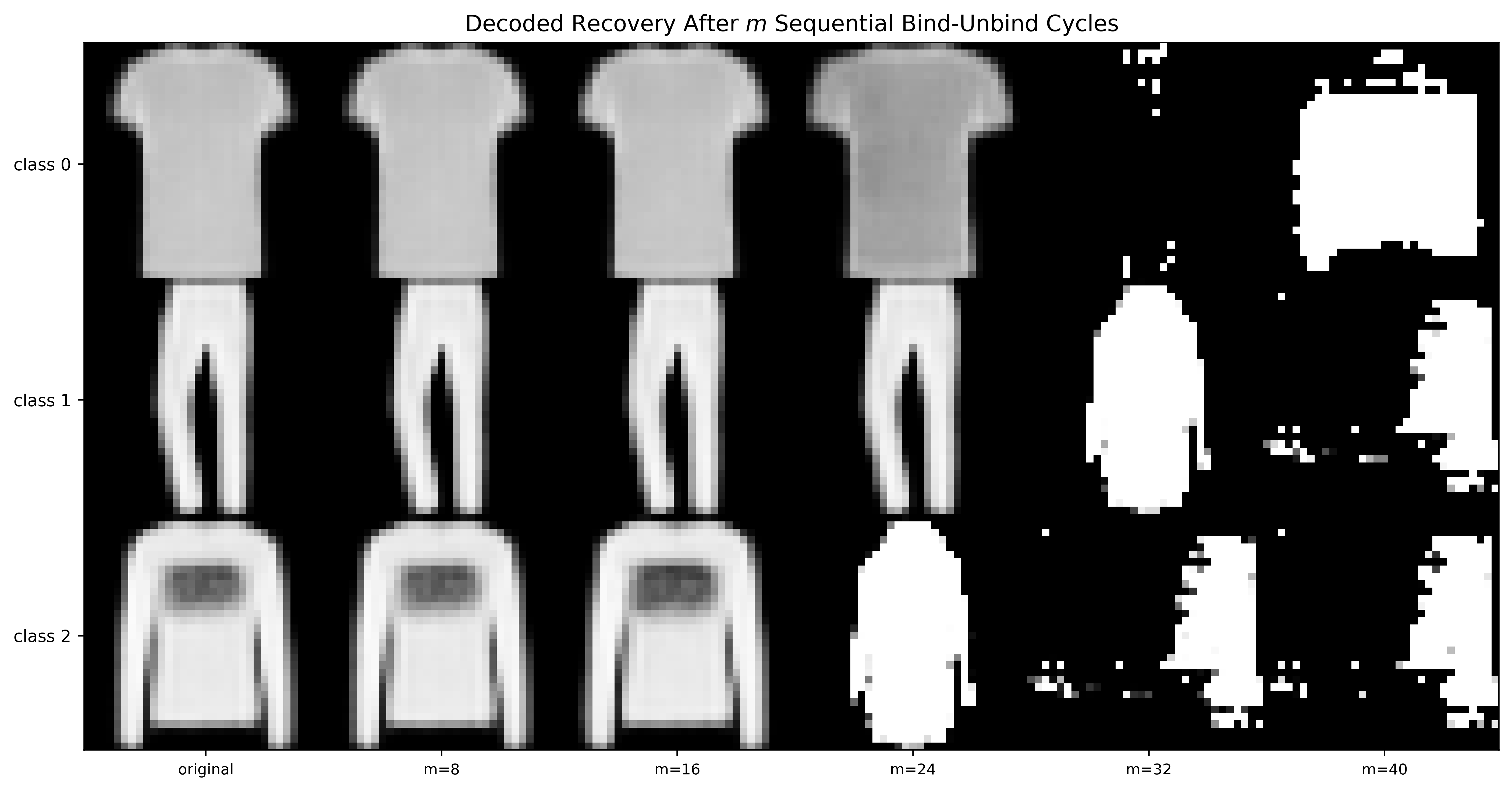}
      \caption{Gaussian L2 ($\dagger$).}
      \label{fig:fashion_rec_gauss_dag}
    \end{subfigure}
    \caption{An image from FashionMNIST is encoded by our model with latent $d=128$, and then bound with $m$ other randomly sampled encoded images from the test set. Each column corresponds to a value of $m$. Decoding is performed with $m$ unbinding operations after the initial $m$ binds. We evaluate the two methods for inverting an HRR proposed in \citet{plate_holographic_1995}, which are equivalent for \textit{unitary} vectors. The phase-only pseudo-inverse is denoted with $*$ and the phase \& magnitude inverse $\dagger$.
} 
      
    \label{fig:fashion_recovery}
  \end{figure}

\begin{figure}[H]
  \centering
  \begin{subfigure}[b]{0.45\linewidth}
    \includegraphics[width=\linewidth]{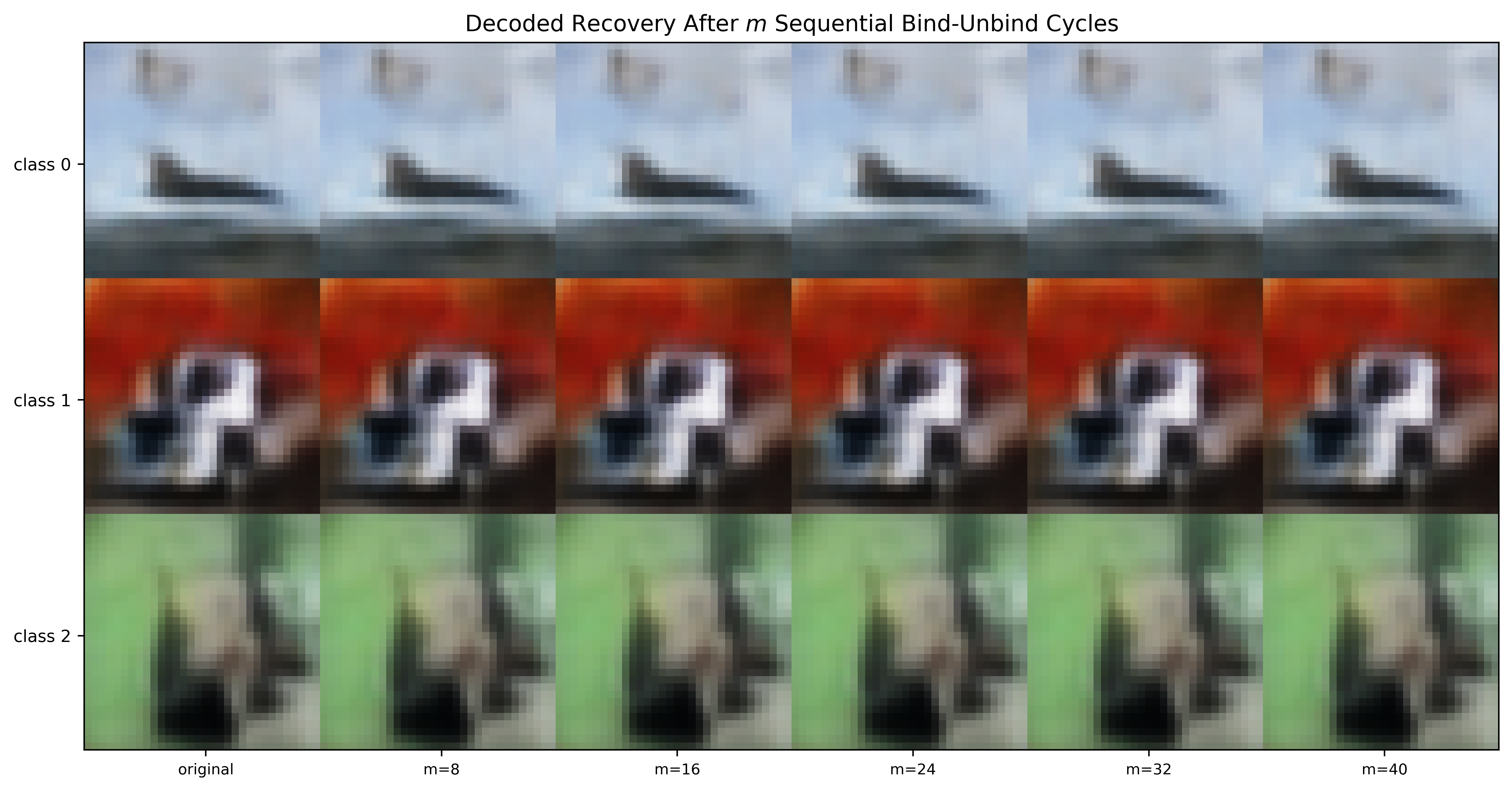}
    \caption{Clifford, involution ($*$).}
    \label{fig:cifar_recovery_clifford_inv}
  \end{subfigure}
  \hfill
  \begin{subfigure}[b]{0.45\linewidth}
    \includegraphics[width=\linewidth]{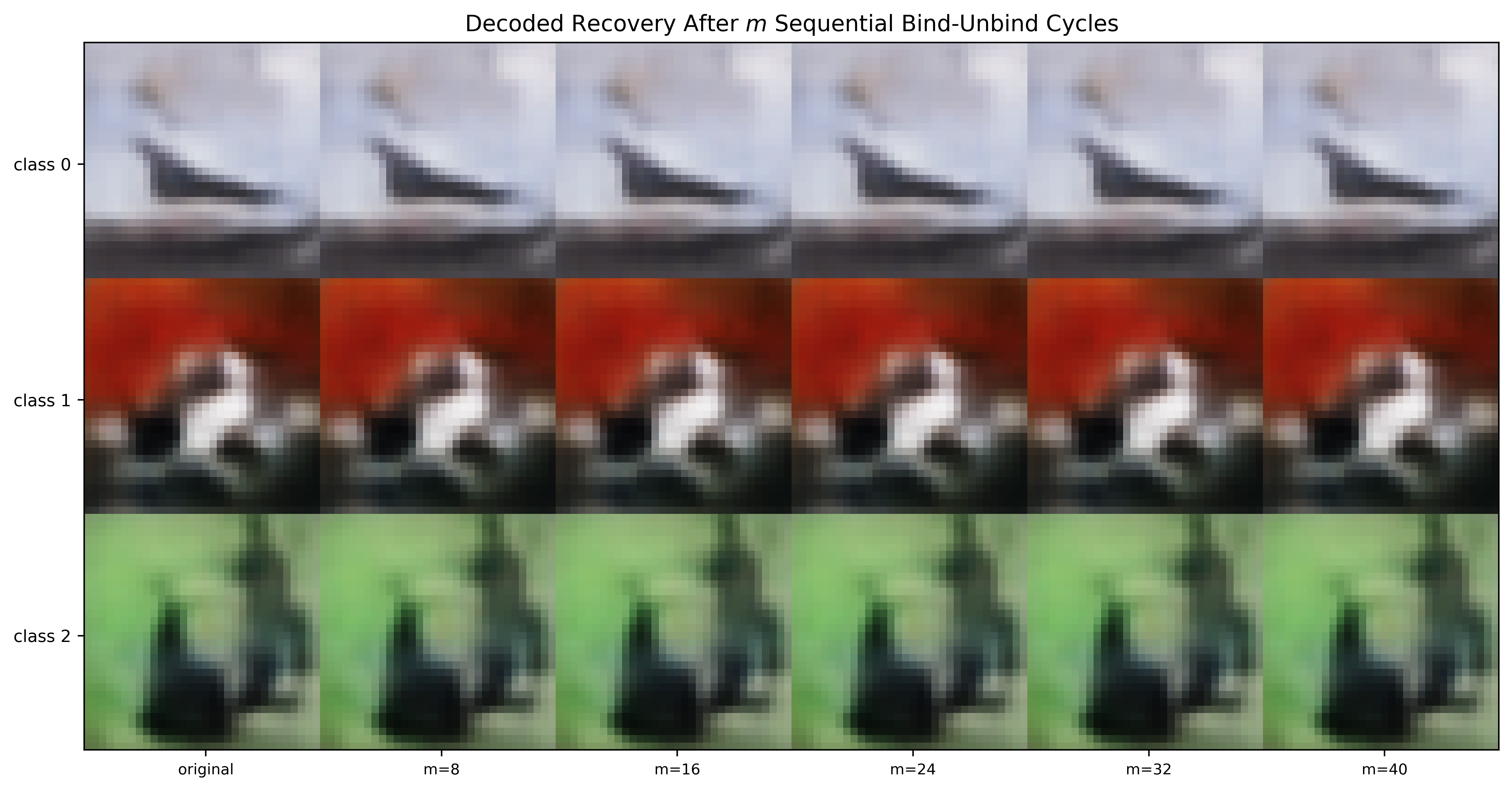}
    \caption{Clifford, ($\dagger$).}
    \label{fig:cifar_recovery_clifford_dagger}
  \end{subfigure}
  \vspace{2pt}
  \begin{subfigure}[b]{0.45\linewidth}
    \includegraphics[width=\linewidth]{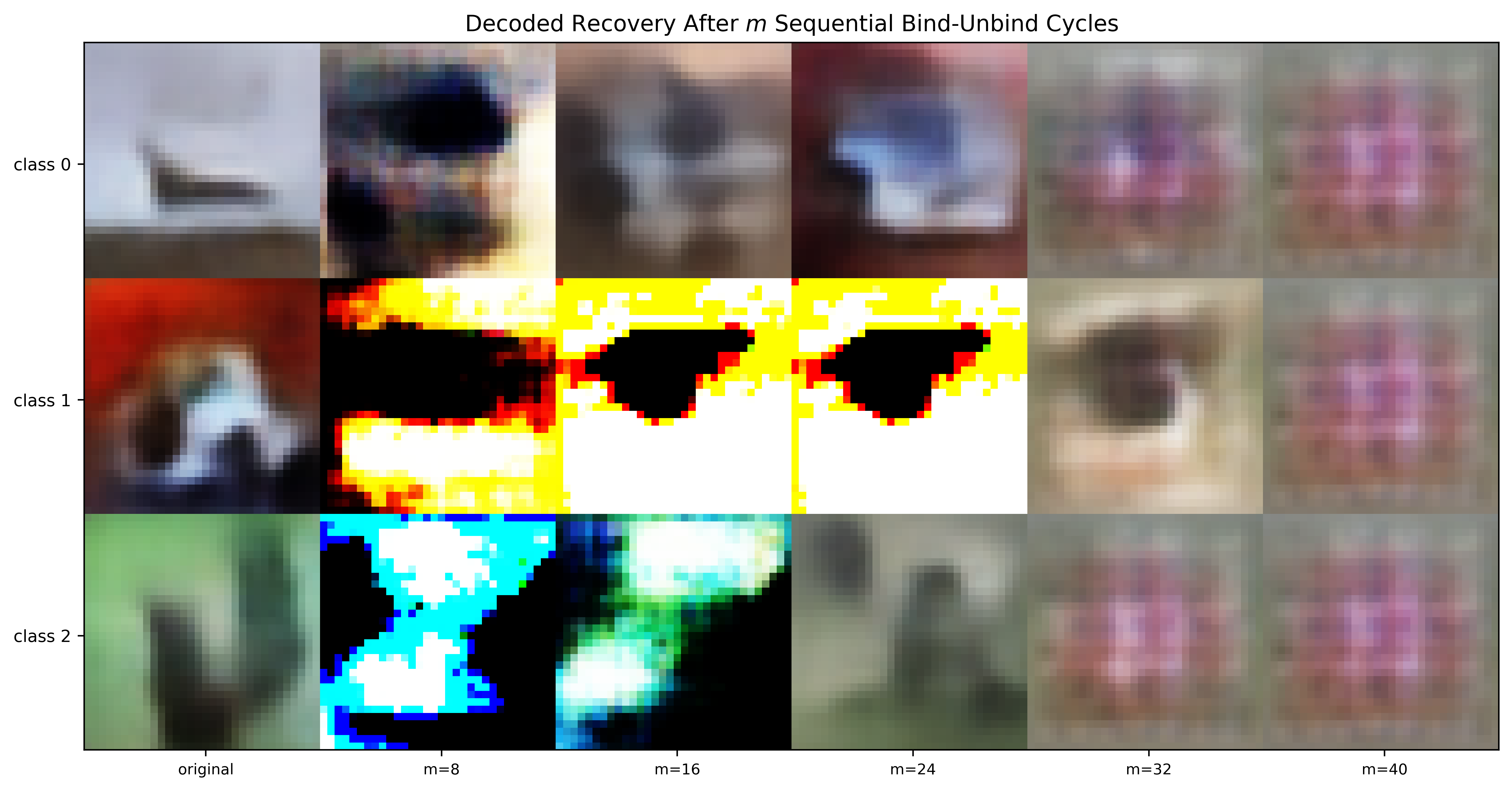}
    \caption{Gaussian, ($*$).}
    \label{fig:cifar_recovery_gaussian_inv}
  \end{subfigure}
  \hfill
  \begin{subfigure}[b]{0.45\linewidth}
    \includegraphics[width=\linewidth]{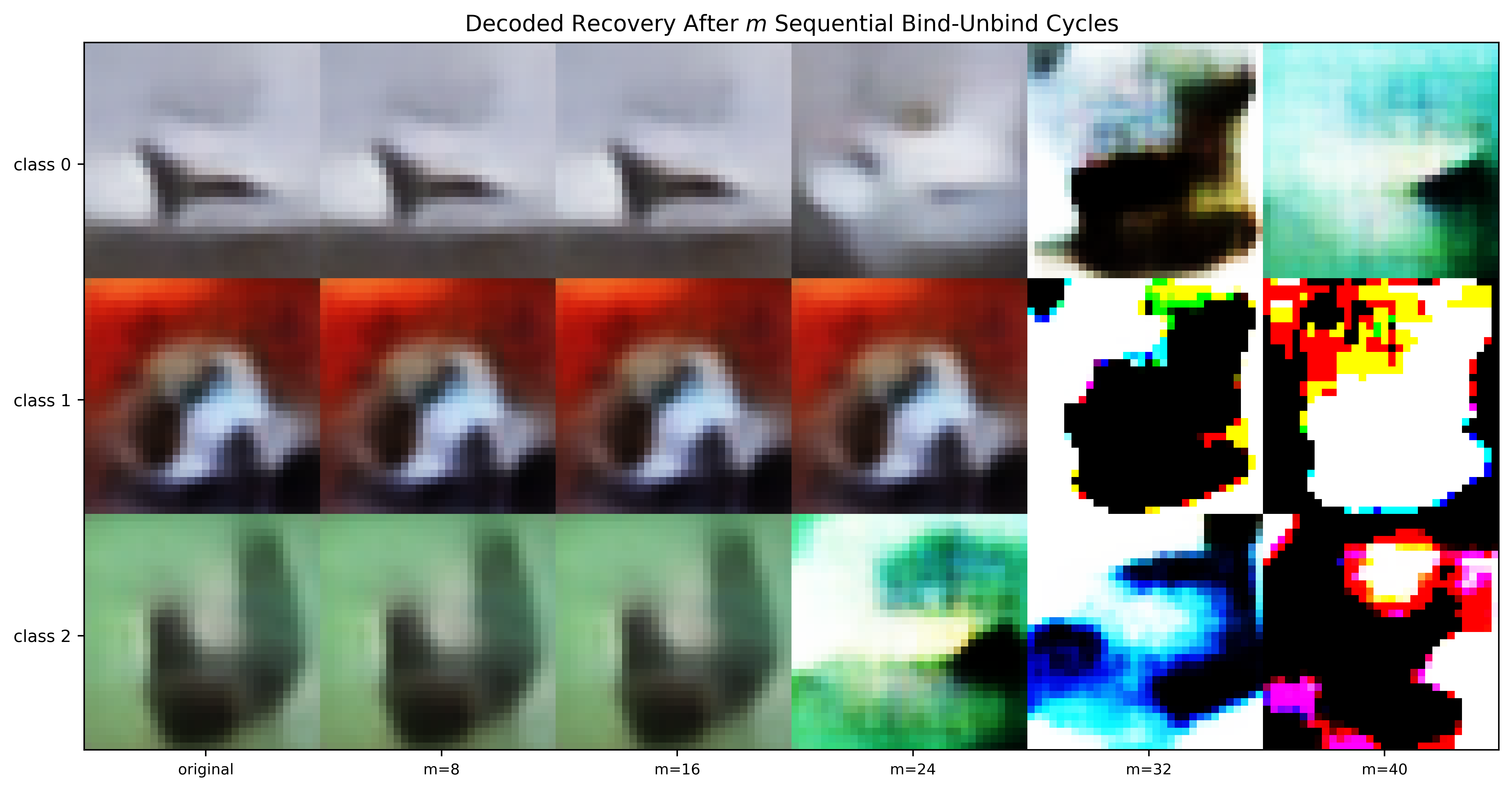}
    \caption{Gaussian, ($\dagger$).}
    \label{fig:cifar_recovery_gaussian_dagger}
  \end{subfigure}
    \caption{Recovery of an image from a compositional VSA encoding, exclusively using our method for atomic vectors. An image from CIFAR-10 is encoded by our model with latent $d=128$, and then bound with $m$ other randomly sampled encoded images from the test set. Gaussian latents lose identity after only a few cycles under $*$, and $\dagger$
    extends recovery slightly by exploiting learned magnitudes before collapsing.}
  \label{fig:cifar_recovery}
\end{figure}

\subsection{VSA Operation Benchmarks}
\label{sec:vsa_benchmarks}

We evaluate the latents encoded by our model for their suitability in use with traditional VSA operations following \citet{Schlegel2020}.
Random unitary vectors sampled via \Cref{eq:unitary_sampling} and canonical HRR random
vectors serve as reference baselines. For all experiments we selected $d=128$ as the latent representation dimensionality.

\begin{figure}[H]
    \centering
    \includegraphics[width=\linewidth]{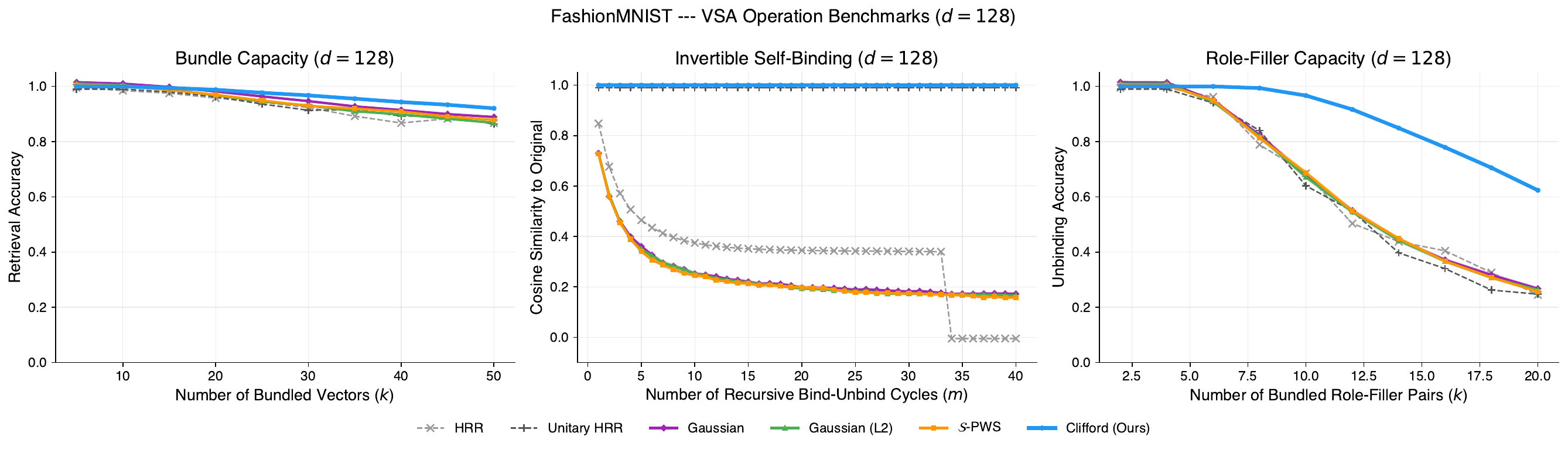}
    \caption{VSA operation benchmarks on FashionMNIST at $d=128$, all distributions overlaid.
      \textit{Left:} bundle capacity. \textit{Center:} role-filler query capacity.
      \textit{Right:} approximate inverse binding depth. Random unitary and canonically
      initialized HRR vectors serve as reference baselines. Shaded bands are the 95\% HDI of the mean (bootstrap, 20k resamples). Results at $d \in \{256, 512, 1024, 2048, 4096\}$ are reported in
      \Cref{appendix:additional_figures}.}
    \label{fig:vsa_fashion_main}
  \end{figure}

\begin{figure}[H]
    \centering 
    \includegraphics[width=\linewidth]{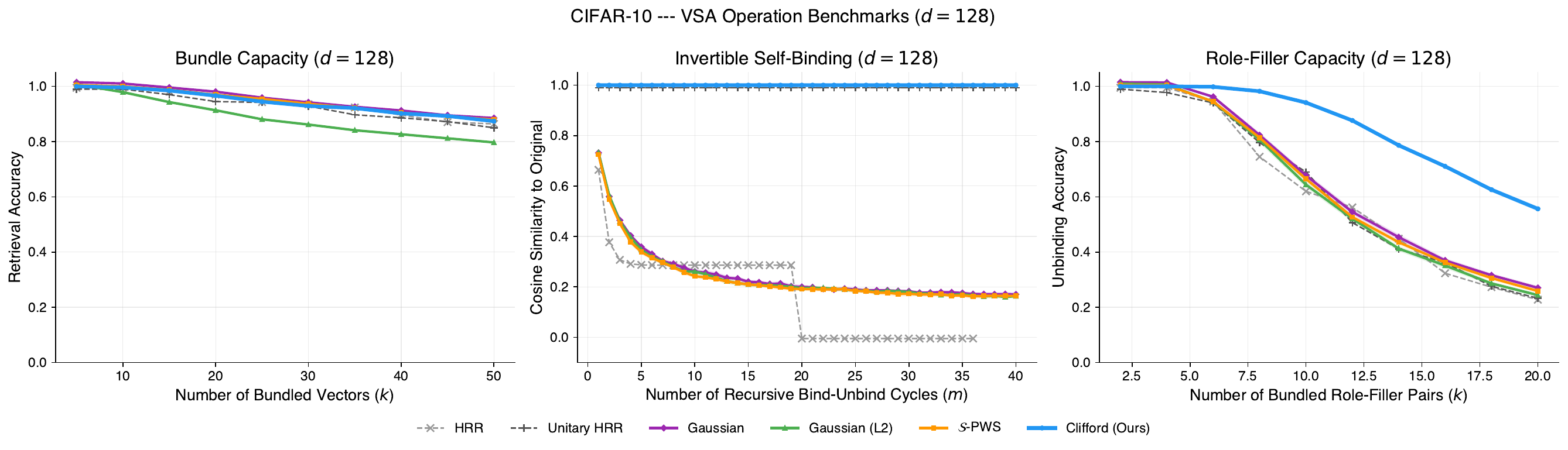}
    \caption{VSA operation benchmarks on CIFAR-10 at $d=128$, all distributions overlaid.
      Panels and baselines as in \Cref{fig:vsa_fashion_main}. Higher-$d$ results in
      \Cref{appendix:additional_figures}.}
    \label{fig:vsa_cifar_main}
  \end{figure}

\paragraph{Bundle capacity.}
Progressively superpose $k$ randomly selected latent vectors via element-wise addition;
measure the fraction recoverable above a cosine similarity threshold. Results shown in the left panel of \cref{fig:vsa_fashion_main} and \cref{fig:vsa_cifar_main}.

\paragraph{Approximate inverse binding depth.}
Recursively bind a vector with $m$ random partners initialized from dataset images encoded
into the latent space, as well as the same image (self-binding). Sequentially unbind using
the $O(d)$ pseudo-inverse and record cosine similarity to the original at each depth $m$.
\Cref{fig:fashion_recovery} and \Cref{fig:cifar_recovery} show a visualization of this, with decoded reconstructions from FashionMNIST and CIFAR-10 after a sequence of $m$ binds and unbinds. Results shown in the center panel of \cref{fig:vsa_fashion_main} and \cref{fig:vsa_cifar_main}.

\paragraph{Role-filler query capacity.}
Bundle $k$ distinct role-filler pairs (roles and fillers are \textit{bound} together) $\sum_i \mathbf{r}_i \circledast \mathbf{f}_i$, then
unbind the bundle by a given role and measure the recovery similarity of the corresponding filler.  Results shown in the right panel of \cref{fig:vsa_fashion_main} and \cref{fig:vsa_cifar_main}.

The above three benchmarks were evaluated using data from the FashionMNIST (\Cref{fig:vsa_fashion_main}) and CIFAR-10 (\Cref{fig:vsa_cifar_main}) datasets embedded using the Clifford-VAE and compared against the same data embedded with the Gaussian, Gaussian-L2, and Power Spherical VAEs, as well as randomly generated HRR and unitary HRR vectors. What we find here is that the performance of the vector representations learned using the Clifford-VAE matches the performance of unitary vectors for bundle capacity and iterated self-binding. Our method outperforms random vectors in the role-filler capacity task for lower dimensions, but as dimensions increase ($d>512$), the performance of all methods converge (see Appendix \ref{appendix:additional_figures}). However, the performance on the iterated self-binding for the other VAE embeddings never matches the performance of the Clifford-VAE, due to not enforcing the unitary constraint.

\citet{kelly2013} evaluated different methods for encoding holographic associative memories, that is to say, creating an association between two vectors by binding them, and then decoding that association by unbinding one of the two elements in the association. In \cref{tab:vsa_comparison} we compare the VSA-encoding cost and VSA-decoding accuracy, as measured by the cosine similarity between the target and decoded vectors. We find that our method behaves as expected, with decoding accuracy comparable to the FHRR/unitary HRR vectors, and encoding cost is that of the HRR binding operator. Should we use the Clifford-VAE in the Fourier space, the complexity would reduce to $O(kn)$.

\begin{table}[H]
\centering
\caption{Comparison of VSAs adapted from \citet{kelly2013}. Our learned
  embeddings match the binding-decoding accuracy of FHRR / unitary HRRs
  \citep[Ch.~4]{plate2003holographic} while supporting unsupervised learning
  from perceptual data. In the Clifford-VAE framework, a vector is produced by a single VAE-encoder forward pass; this cost is not accounted for in the VSA-encoding of associative memories.}
\label{tab:vsa_comparison}
\resizebox{\textwidth}{!}{%
\begin{tabular}{lllcc}
\toprule
VSA & Representation & Encoding & Encoding cost & Decoding acc.\ \\
\midrule
Tensor representations \citep{smolensky1990}
  & rank-$k$ tensor of $n^k$ reals & tensor product & $O(n^k)$ & $1.00$ \\
HRR \citep{plate2003holographic}
  & vector of $n$ reals & circular convolution $\circledast$ & $O(kn \log n)$ & $0.71$ \\
Frequency-domain HRR \citep{plate2003holographic}
  & vector of $n$ unit-circle complex values & element-wise mult.\ $\circ$ & $O(kn)$ & $1.00$ \\
Binary spatter codes \citep{Kanerva1996BinarySO}
  & vector of $n$ binary values & bit-wise XOR & $O(kn)$ & $1.00$ \\
MAP coding \citep{gayler-2004}
  & vector of $n$ values in $\{-1,+1\}$ & element-wise mult.\ $\circ$ & $O(kn)$ & $1.00 / 0.58$ \\
Square-matrix repr.\ ~\citep{kelly2010}
  & $\sqrt{n} \times \sqrt{n}$ real matrix & matrix multiplication & $O(kn^{3/2})$ & $0.70$ \\
CHARM~\citep{eich1982}
  & vector of $n$ reals & truncated aperiodic conv.\ & $O(kn \log n)$ & $0.70$ \\
\midrule
\textbf{Clifford-VAE (ours)}
  & vector of $n$ unit-circle complex values & circular convolution $\circledast$ & $O(kn \log n)$ & $1.00$ \\
\bottomrule
\end{tabular}%
}
\end{table}

%%%%%%%%%%%%%%%%%%%%%%%%%%%%%%%%%%%%%%%%%%%%%%%%%%%%%%%
\section{Discussion}
%%%%%%%%%%%%%%%%%%%%%%%%%%%%%%%%%%%%%%%%%%%%%%%%%%%%%%%

We have formulated a Variational Autoencoder that replaces the Gaussian prior with a
distribution on the $(d-1)$-dimensional Clifford torus $(\mathcal{S}^1)^{d-1}$, which serves to strictly constrain 
latent codes to be learned with Fourier coefficients of unit magnitude. Our approach provides an unsupervised method to translate unstructured data into representations that are compatible with the HRR algebra. The representations learned by our Clifford-VAE implementation match or surpass unitary/FHRR accuracy on recursive bind/unbind, role-filler recovery, and bundle capacity tests. 
When used as a feature embedding for downstream classification tasks, we find that the hyperspherical constraints of the Clifford-VAE do not preclude competitive performance. On CIFAR-10, our model improves over the strongest baseline ($\mathcal{N}$) in all evaluated dimensions and sample sizes for supervised learning (\textit{k-NN}) by $29\%$ on average (ranging from $+13.4\%$ at $d=4096$ to
  $+48.4\%$ at $d=128$). The gains in accuracy observed are generally most prevalent at the regime with $n_{\ell}=100$ labeled samples from which to fit a $k$-NN model with $k=5$, with additional labels closing the gap between our method and baselines.

On the simpler FashionMNIST dataset, we observe best-in-class performance for $n_{\ell}=100$, with a modest improvement of $\approx4.1\%$. For higher labeling budgets, our method is still highly competitive (within $\pm 2\%$).

Our results on MNIST under the conditions tested by \citet{davidson2018} show that Clifford-VAE has comparable or better performance relative to the standard VAE. Hyperspherical variants show declining performance as dimensionality increases, further motivating our formulation of the Clifford torus as an alternative prior. As $n_{\ell}$, the number of samples that the supervised learning model is fit to, increases, we see the performance of all models improve, but the general trends for the $n_{\ell}=100$ cases are preserved. In all cases, the gap in performance that our model shows relative to the strongest baseline narrows as $d$ increases, and at $d=128$, our Clifford implementation of \citet{davidson2018}'s VAE architecture outperforms all other choices of prior, which is advantageous given our motivation to train VAEs that produce high-dimensional representations that are compatible with the HRR algebra.

In the case of image reconstruction, we see that the Clifford-VAE model has worse performance than the standard VAE, as measured by FID score. But the proposed method is more robust to dimensionality increase than the Power Spherical VAE. All models except the L2-normalized Gaussian showed reconstruction quality that degraded with increasing $d$, reflecting the lack of tuned 
  regularization as $d$ was increased. However, that advantage was not preserved when the latent representation was integrated with iterated VSA operations. This suggests that while our embedding may lose some reconstruction fidelity, it is more robust to noise that results from algebraic operations.

While the proposed Clifford-VAE can act as a reasonable encoding-decoding structure, the utility is greatest in the context of VSAs. \citet{kelly2013} noted the limitations of relying on randomly-generated atomic symbols in HRR-based
cognitive models, elaborating upon 
\citet{chalmers1992highlevel}. With random vector generation, conceptual processes are rendered independent of perceptual processes. Even random projections like fractional power encoding assume some independence between perception and cognition, or at the very least may pay some price for choosing suboptimal representations. 

As our learned representations are semantically meaningful, decodable back into the perceptual domain, and well-behaved
under the HRR algebra, our method for learning VSA embeddings could thus serve as a crucial component of various HRR-based cognitive models, as well as implementations of hyperdimensional computing (HDC) systems in practical engineering settings.

%% embeddings, sample efficiency, mention HRRformer and similar ideas 

\paragraph{Limitations.}
The Clifford torus restricts the encoder to representations whose Fourier coefficients all
have unit magnitude, potentially limiting expressivity for tasks where varying magnitudes can carry useful
information. The conjugate symmetric constraint in our sampling methodology doubles decoder input dimension ($2d$ vs.\ $d$),
increasing parameter count, although this does not augment the input with any useful information relative to other variants, but rather simply ensures that our samples are real-valued. Our experiments cover MNIST, FashionMNIST, and CIFAR-10, given our focus on supporting learnable neuro-symbolic representations rather than investigating the Clifford-VAE or alternative methods leveraging KL-regularization for more sophisticated downstream machine learning tasks, \emph{i.e.}, VSA computations more complicated than the benchmark tasks.
\paragraph{Future work.}
A Clifford-VAE encoder could serve as a differentiable `perceptual front-end' for HRR-based
cognitive architectures. One could imagine constraining the learned representations by their utility in VSA-defined functions, in addition to the Clifford torus constraint.

%%%%%%%%%%%%%%%%%%%%%%%%%%%%%%%%%%%%%%%%%%%%%%%%%%%%%%%

Our method attempts to unify the HRR VSA,
which retain algebraic guarantees for the manipulation of distributed representations, and the
probabilistic deep learning of distributed representations, akin to how
\citet{bengio03neural} introduced the ability to learn neural probabilistic representations of words. In his pioneering work, the joint probability function of word sequences is expressed in terms of the \textit{feature vectors} of the words; thus, both the parameters of the function and the feature vectors themselves are learned. Analogously, our Clifford-VAE can be seen as a method for encoding arbitrary data into HRR embeddings, enabling operations that exploit the context-sensitive representations and their similarity metric for predictive or generative tasks. 

Alternatively, viewed through the lens of the binding problem as surveyed in \citet{binding-greff-2020}, our work can thus be viewed as addressing the first challenge, representation, enabling future efforts to solve the subsequent stages that effective neuro-symbolic systems may implement: namely, learning to disambiguate between levels of hierarchy (segregation), and learning to compose the HRR representations into reusable and interpretable structures applicable for various downstream tasks.

\section{Methods}
%%%%%%%%%%%%%%%%%%%%%%%%%%%%%%%%%%%%%%%%%%%%%%%%%%%%%%%

\subsection{Variational Autoencoder Training Objective}

\begin{equation}
\mathcal{L}_{\text{ELBO}} =
  \mathbb{E}_{q_\phi}[\log p_\theta(\mathbf{x}|\mathbf{z})]
  - \beta \cdot \text{KL}[q_\phi \| p]
\end{equation}

The ELBO is optimized with Adam for up to
  500 epochs with early stopping on the total loss. For MNIST, following
  \citet{davidson2018}, we use a BCE reconstruction loss on dynamically binarized
  inputs and a linear KL warm-up, annealing $\beta$ from $0$ to $1$ over the first
  100 epochs. For FashionMNIST and CIFAR-10 we use an L1 reconstruction loss and,
  after the same 100-epoch warm-up, a cyclical (triangular) $\beta$ schedule of period
  250 epochs oscillating between $\beta_{\min}=0.1$ and $\beta_{\max}=1.0$, following \citet{fu2019}. Full configurations are detailed in \Cref{tab:hyperparams}.

  \begin{table}[H]
  \centering
  \caption{Training hyperparameters:
  ``cyclical'' denotes a triangular $\beta$ schedule oscillating between
  $\beta_{\min}$ and $\beta_{\max}$ with the given period after warm-up;
  ``warm-up'' denotes a single linear ramp of $\beta$ from $0$ to $1$.}
  \label{tab:hyperparams}
  \resizebox{\textwidth}{!}{%
  \begin{tabular}{lllllllll}
  \toprule  
  Dataset & Arch. & Optim. & Batch & LR & Recon. & $\beta$ schedule & Epochs (patience) & Trials \\
  \midrule
  MNIST        & MLP & Adam  & 128 & $10^{-3}$         & BCE (binarized) & warm-up, 100 ep & 500 (50) & 20 \\
  FashionMNIST & CNN & AdamW & 256 & $3\times10^{-4}$  & L1 & cyclical $[0.1,1.0]$, period 250 & 500 (50) & 30 \\
  CIFAR-10     & CNN & AdamW & 256 & $3\times10^{-4}$  & L1 & cyclical $[0.1,1.0]$, period 250 & 500 (50) & 30 \\
  \bottomrule
  \end{tabular}%
  }
  \vspace{2pt}
  \footnotesize All runs use a 100-epoch KL warm-up before the schedule above and
  early stopping on the total loss.
  \end{table}

We evaluate latent dimensions
  $d\in\{2,5,10,20,40,128\}$ on MNIST (extending \citet{davidson2018} past $d=40$), and   $d\in\{128,256,512,1024,2048,4096\}$ on FashionMNIST and CIFAR-10.
  We note that while vMF numbers fall short of those originally reported in \citet{davidson2018}, we elected to instead use the more tractable Power Spherical distribution
  \citep{decao2020} as our main hyperspherical baseline, retaining vMF on MNIST only for direct
  comparison with \citet{davidson2018}. Furthermore, the numbers achieved in the past necessitated the use of careful hyperparameter tuning, float64 precision during sampling, and a slower learning schedule to avoid posterior collapse, which we found were not necessary with use of the more recent Power Spherical implementation for comparison.\footnote{T.\ R.\ Davidson, personal
  communication, July 2025.} 
 Results are reported as the mean with its 95\% HDI; see \Cref{appendix:hdi} for details.

\subsubsection{Reconstruction Losses}

Following \citet{davidson2018}, we replicate the MNIST experiments with BCE on binarized data.
\textbf{BCE} (MNIST):
\begin{equation}
\mathcal{L}_{\text{recon}} = -\frac{1}{B}\sum_{i=1}^B \sum_{j=1}^D
  \left[ x_{ij} \log \hat{x}_{ij} + (1-x_{ij}) \log(1-\hat{x}_{ij}) \right]
\end{equation}

For FashionMNIST and CIFAR-10, we optimize for the L1 reconstruction loss as part of our ELBO.
\textbf{L1} (FashionMNIST, CIFAR-10):
\begin{equation}
\mathcal{L}_{\text{recon}} = \frac{1}{B}\sum_{i=1}^B \|\mathbf{x}_i - \hat{\mathbf{x}}_i\|_1
\end{equation}

\subsubsection{KL Annealing}
Cyclical $\beta$-annealing, an extension of the $\beta$-VAE framework,
has been shown to mitigate the phenomenon of KL vanishing by cycling the $\beta$ constant for a given number of epochs between $\lbrack\beta_{\min}, \beta_{\max}\rbrack$:
\begin{equation}
\beta_{\text{cyclic}}(e) =
  \beta_{\min} + (\beta_{\max} - \beta_{\min}) \cdot \tau(e)
\end{equation}
where $\tau(e)$ is a triangular schedule with period $E_{\text{cycle}}$.

\subsubsection{KL Divergence Derivation}

\label{sec:kl_divergence}
The variational posterior on the Clifford torus is a product of independent circular distributions: the KL divergence against the uniform prior on $\mathcal{CT}^{d-1}$ factorizes into a
  sum of per-circle terms \citep{increasingexpressivitydavidson}:
  \begin{equation}
  \text{KL}[q_\phi(\mathbf{z}|\mathbf{x}) \| p(\mathbf{z})] =
    -H[q_\phi] + (d-1)\log(2\pi)
  \end{equation}
where $H[q_\phi]$ is the differential entropy of the product posterior, computed in closed form (\Cref{tab:entropy_summary}), and the uniform prior on $\mathcal{CT}^{d-1}$ has entropy $(d-1)\log(2\pi)$.

We initially model each circular factor with a von Mises distribution, which samples from 
  $\mathcal{S}^1$ with mean direction $\mu_i$ and concentration $\kappa_i$. Taking the $d-1$ free
  angles to be independent, the posterior density and its differential entropy factorize over the
  circles as
\begin{equation}\label{eq:vm-pdf}
p(\boldsymbol{\theta}|\boldsymbol{\mu}, \boldsymbol{\kappa}) =
  \prod_{i=1}^{d-1} \frac{e^{\kappa_i \cos(\theta_i - \mu_i)}}{2\pi I_0(\kappa_i)}.
\end{equation}
Substituting \cref{eq:vm-pdf} into equation 4 of \citet{increasingexpressivitydavidson} yields the differential entropy of this distribution, $H[\text{CT-vM}]$:
\begin{equation}
H[\text{CT-vM}] = \sum_{i=1}^{d-1}
  \left[\log(2\pi I_0(\kappa_i)) - \kappa_i \frac{I_1(\kappa_i)}{I_0(\kappa_i)}\right].
\end{equation}

Alternatively, each circle may be sampled from a Power Spherical distribution~\citep{decao2020},
  which is defined directly on $\mathcal{S}^1$ and admits a closed-form, numerically stable
  reparameterization. Representing each angle as a unit vector
  $\mathbf{v}_i = (\cos\theta_i, \sin\theta_i) \in \mathcal{S}^1$, the per-factor density and entropy are:
\begin{equation}\label{eq:ps-pdf}
p(\mathbf{v}_i|\boldsymbol{\mu}_i, \kappa_i) =
  N_X(\kappa_i, 2)^{-1}
  (1 + \boldsymbol{\mu}_i^T\mathbf{v}_i)^{\kappa_i},
  \qquad
  N_X(\kappa_i, 2) = \frac{2\pi\Gamma(\kappa_i)}{\Gamma(\frac{1}{2} + \kappa_i)}
\end{equation}
where $N_X(\kappa, d)$ is the Power Spherical normalizing constant of \citet{decao2020}, here specialized to the circle ($d=2$).
Substituting \cref{eq:ps-pdf} into equation 4 of \citet{increasingexpressivitydavidson} yields the differential entropy for the Power Spherical-sampled Clifford Torus distribution:
\begin{equation}
H[\text{CT-PS}] = \sum_{i=1}^{d-1}
  \left[\log N_X(\kappa_i, 2)
    - \kappa_i\bigl(\log 2 + \psi(\alpha_i) - \psi(\alpha_i + \beta_i)\bigr)\right].
\end{equation}
where $\alpha_i = \frac{1}{2} + \kappa_i$, $\beta_i = \frac{1}{2}$, and $\psi(\cdot)$ is
the digamma function.

\begin{table}[H]
\centering
\caption{Differential entropy of distributions.}
\label{tab:entropy_summary}
\begin{tabular}{ll}
\toprule
\textbf{Distribution} & \textbf{Entropy $H$} \\
\midrule
von Mises & $\log(2\pi I_0(\kappa)) - \kappa \frac{I_1(\kappa)}{I_0(\kappa)}$ \\[4pt]
Power Spherical ($\mathcal{S}^{d-1}$) \citep{decao2020} & $\log N_X(\kappa,d) - \kappa(\log 2 + \psi(\alpha) - \psi(\alpha+\beta))$ \\
 & $\alpha = \frac{d-1}{2} + \kappa$, $\beta = \frac{d-1}{2}$ \\[4pt]
CT-vM & $\sum_{i=1}^{d-1} \left[\log(2\pi I_0(\kappa_i)) - \kappa_i \frac{I_1(\kappa_i)}{I_0(\kappa_i)}\right]$ \\[4pt]
CT-PS & $\sum_{i=1}^{d-1} \left[\log N_X(\kappa_i,2) - \kappa_i(\log 2 + \psi(\alpha_i) - \psi(\alpha_i+\beta_i))\right]$ \\[4pt]
Uniform on $\mathcal{CT}^{d-1}$ & $(d-1)\log(2\pi)$ \\
\bottomrule
\end{tabular}
\end{table}

\subsection{Sampling from the Clifford Torus}
\label{appendix:ct_distributions}

A Clifford torus is a flat torus embedded in the 3-sphere $\mathcal{S}^3 \subset \mathbb{R}^4$,
defined as the Cartesian product of two circles. The $(d-1)$-dimensional generalization
$\mathcal{CT}^{d-1}$ is the product manifold $(\mathcal{S}^1)^{d-1}$ embedded in $\mathbb{R}^{2(d-1)}$:
\begin{equation}
\mathcal{CT}^{d-1} = \left\{
  [(\cos\theta_1, \sin\theta_1), \ldots, (\cos\theta_{d-1}, \sin\theta_{d-1})]
  \in \mathbb{R}^{2(d-1)}
  \;\middle|\;
  \theta_i \in [0, 2\pi)
\right\}.
\end{equation}

Standard Gaussian VAEs parameterize the latent distribution as
$q_\phi(\mathbf{z}|\mathbf{x}) = \mathcal{N}(\boldsymbol{\mu}_\phi(\mathbf{x}),
  \mathrm{diag}(\boldsymbol{\sigma}^2_\phi(\mathbf{x})))$. The real and imaginary parts of each non-DC Fourier coefficient are i.i.d.\
  $\mathcal{N}(0,\sigma^2 d/2)$, thus, magnitudes follow
  $|\mathcal{F}\{\mathbf{z}\}_k| \sim \mathrm{Rayleigh}(\sigma\sqrt{d/2})$, with nonzero
  variance.  
  Hence while L2-normalized Gaussian and Hyperspherical VAEs enforce
$\|\mathbf{z}\| = 1$, the necessary constraints on individual Fourier coefficient magnitudes for effective binding and unbinding with circular convolution are not ensured.

In order to satisfy these constraints by geometric construction, we choose to learn $\mu_i$ variationally, sampling:
$$\mathbf{v}_i \sim \mathrm{PowerSpherical}\bigl((\cos\mu_i, \sin\mu_i),\,\kappa_i\bigr),
  \quad \forall i \in \{1, \ldots, d-1\}$$
and recover the angle via $\theta_i = \arctan2(v_{i,1}, v_{i,0})$. Alternatively, a circular distribution can be utilized to sample angles directly, and recent work has explored more efficient implementations of the von Mises/von Mises-Fisher~\citep{Pakman_2025,kim-2021}. Further alternatives such as the Spherical Cauchy distribution, have also been proposed~\citep{sablica2025spcauchy}.

Taking the inverse Fourier transform $\mathcal{F}^{-1}$ of $e^{i\theta}$ produces complex-valued output, incompatible with real-valued
decoder layers. In order to ensure real outputs, the frequency vector must be Hermitian:
$h_k = \overline{h_{d-k}}$. Given
$\boldsymbol{\theta} = (\theta_1, \ldots, \theta_{m-1})$, we construct
$\mathbf{h} \in \mathbb{C}^{d}$ with $d = 2m$:
\begin{equation}
h_k = \begin{cases}
0 & k = 0 \\
e^{i\theta_k} & k \in \{1, \ldots, m-1\} \\
0 & k = m \\
e^{-i\theta_{2m-k}} & k \in \{m+1, \ldots, 2m-1\}
\end{cases}
\end{equation}

The inverse Fourier transform then produces $\mathbf{x} \in \mathbb{R}^{2m}$ with
$|\hat{\mathbf{x}}_k| = 1$ exactly for all non-DC frequencies, and
$\|\mathbf{x}\|_2^2 = \frac{m-1}{m} \to 1$ as $m \to \infty$.

% \subsubsection{CT von Mises (CT-vM)}

% \subsubsection{CT Power Spherical (CT-PS)}

% \subsubsection{Sampling Procedures}
Each subspace of our distribution is sampled with a reparameterizable scheme such that gradients flow to the posterior
  parameters $(\mu_i,\kappa_i)$. The Power Spherical admits a closed-form reparameterization via its
  inverse CDF~\citep{decao2020} (Algorithm \ref{alg:ct-ps}), and the von Mises-Fisher is sampled with the rejection-sampling
  reparameterization of \citet{davidson2018} (Algorithm \ref{alg:ct-vm}).

\begin{algorithm}
\caption{CT-vM: Sampling points on the Clifford torus using the von Mises distribution.}\label{alg:ct-vm}
\begin{algorithmic}
\State $\theta_i \sim \mathrm{vonMises}(\mu_i, \kappa_i)$ for $i=1,\ldots,d-1$
\State $\boldsymbol{\theta}_s \gets [0, \theta_1, \ldots, \theta_{d-1}, 0, -\theta_{d-1}, \ldots, -\theta_1]$
\State $\mathbf{x} = \mathcal{F}^{-1}\left\{e^{i\boldsymbol{\theta}_s}\right\}$
\end{algorithmic}
\end{algorithm}

% \paragraph{CT-vM}
% \begin{enumerate}
%   \item Sample 
%   \item 
%   \item Compute 
% \end{enumerate}

\begin{algorithm}
\caption{CT-PS: Sampling points on the Clifford torus using the Power Spherical distribution to select individual complex elements.}\label{alg:ct-ps}
\begin{algorithmic}
\State $\mathbf{v}_i \sim \mathrm{PowerSpherical}((\cos\mu_i, \sin\mu_i), \kappa_i)$ on $\mathcal{S}^1$ for $i=1,\ldots,d-1$
\State $\theta_i = \arctan2(v_{i,1}, v_{i,0})$  for $i=1,\ldots,d-1$
\State $\boldsymbol{\theta}_s \gets [0, \theta_1, \ldots, \theta_{d-1}, 0, -\theta_{d-1}, \ldots, -\theta_1]$
\State $\mathbf{x} = \mathcal{F}^{-1}\left\{e^{i\boldsymbol{\theta}_s}\right\}$
\end{algorithmic}
\end{algorithm}

% \paragraph{CT-PS}
% \begin{enumerate}
%   \item Sample 
%   \item Set 
%   \item Compute 
% \end{enumerate}

\begin{figure}[H]
  \centering
  \includegraphics[width=\linewidth]{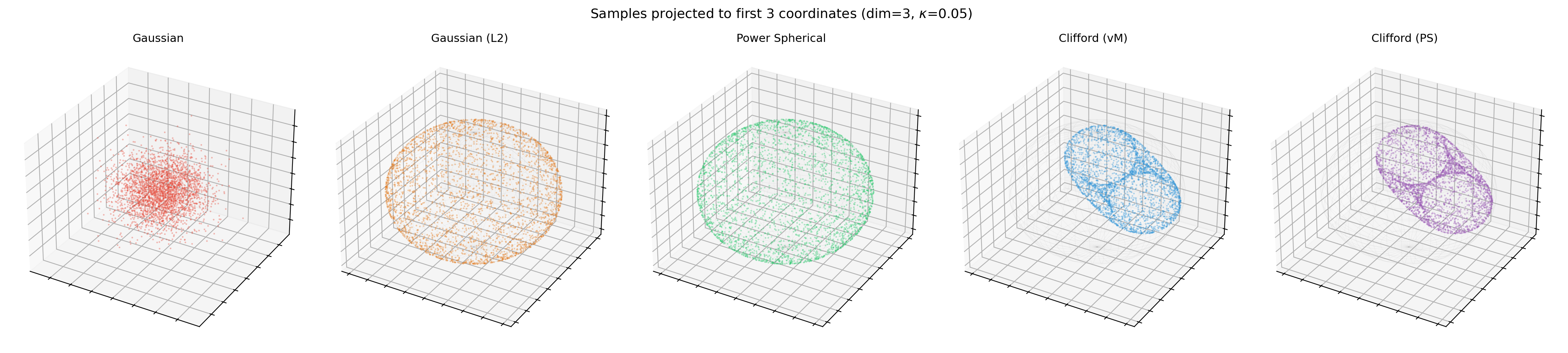}
  \caption{3D projections of 3000 samples ($d=3$, $\kappa=0.05$) from Gaussian distributions
    (both with and without L2 normalization), a Power Spherical distribution, and our two
    Clifford constructions (CT-vM and CT-PS).}
  \label{fig:distribution_samples}
\end{figure}

\Cref{fig:distribution_samples} visualizes samples from each prior in three dimensions: the Gaussian
latents fill the interior of the space, while the L2-normalized Gaussian variant and the Power Spherical
concentrate on the sphere's surface. In contrast, our distributions place mass on the
product-of-circles torus.

\subsection{Architecture}
On MNIST, we follow the implementation of \citet{kingma2014} as used by \citet{davidson2018}, specifically training only the M1 model rather than M1+M2. M1 is trained as an MLP with 
  encoder $784 \to 256 \to 128$, and the decoder mirroring it, with the final layer outputting logits through a sigmoid to parameterize the per-pixel Bernoulli mean $\hat{x}_{ij}$ in the BCE. 
  For our experiments on FashionMNIST and CIFAR-10, we train a convolutional neural network with four residual blocks (Conv2d + LeakyReLU($0.2$) +
  skip connection) and channels $[64,128,256,512]$, mapping a $32\times32$ input to a $2\times2$
  spatial map (flattened dimension $2048$); the decoder is the transposed mirror with a $\tanh$ activation. The Clifford decoder
receives $2d$-dimensional input due to the conjugate symmetry constraint; other decoders receive
$d$.

\subsection{Supervised Learning Protocol (k-Nearest Neighbor)}
  \label{sec:methods_knn}
   Each VAE is
  first trained fully unsupervised; its encoder (recognition network
  $q_\phi(\mathbf{z}\mid\mathbf{x})$) is then \emph{frozen} and used as a fixed feature
  extractor. For each image we take the latent representation
  produced by the trained encoder as its feature vector, and fit a $k$-nearest-neighbor classifier with $k=5$ neighbors and a cosine distance metric (\textit{Euclidean} for Gaussian variants $\mathcal{N}$
  and $\mathcal{N}_{L2}$). 
  We evaluate over label budgets $n_{\ell}\in\{100,600,1000\}$ and report top-1 accuracy on the held-out test set, averaged over
  $n=20$ independent trials per dimension on MNIST and $n=30$ on FashionMNIST and CIFAR-10.
  The $n_{\ell}$ labeled exemplars are drawn at random and accuracy is
  measured on the full test set.

\subsection{Statistical reporting}
  \label{appendix:hdi}
  All results and figures report the mean over runs together with the 95\% highest-density
  interval (HDI) of the mean, estimated by bootstrap: from the $N$ per-run values in a cell we
  draw $N$ values with replacement, take their mean, and repeat $20{,}000$ times; the 95\% HDI is
  the narrowest interval covering 95\% of that bootstrap distribution. 
  The best mean is \textbf{\underline{bold and underlined}} when its HDI is disjoint from every other entry's (a clear winner);
  when the best mean is statistically tied with one or more entries (their HDIs overlap), those tied entries are
  \underline{underlined}.
  
\section*{Funding Declaration}
  The authors received no funding for this work.

\newpage

%%%%%%%%%%%%%%%%%%%%%%%%%%%%%%%%%%%%%%%%%%%%%%%%%%%%%%%
\bibliographystyle{plainnat}
\bibliography{references}

\newpage
\appendix

\section{Additional Figures}
  \label{appendix:additional_figures}

  \subsection{FashionMNIST VSA benchmarks}
  \begin{figure}[H]
    \centering
    \includegraphics[width=\linewidth]{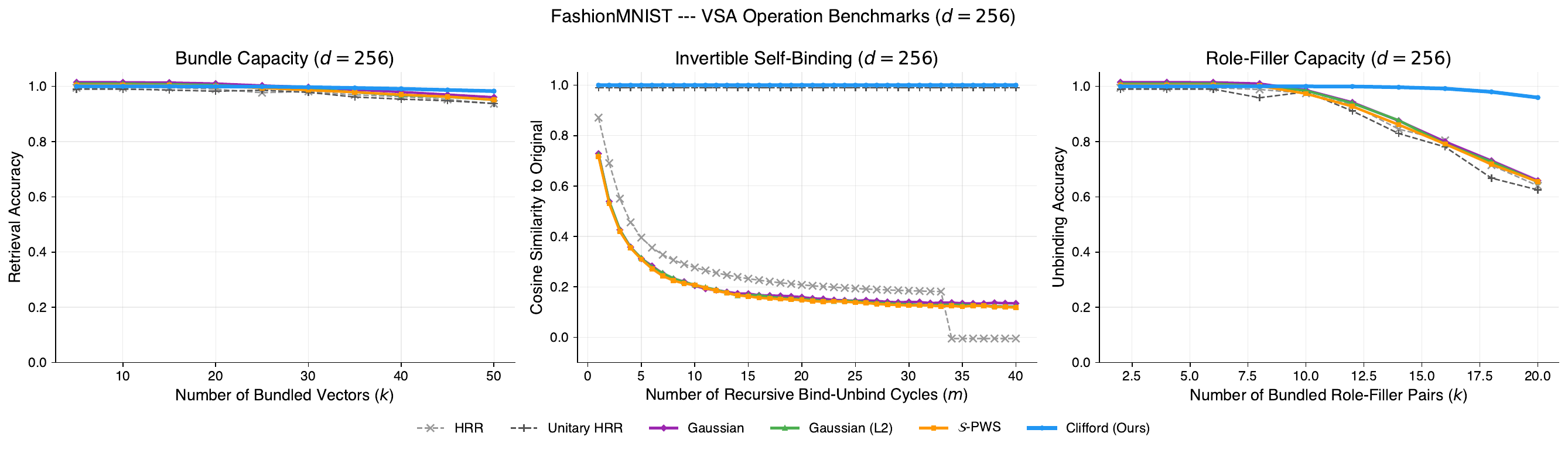}
    \caption{FashionMNIST $d=256$.}
  \end{figure}
  \begin{figure}[H]   
    \centering
    \includegraphics[width=\linewidth]{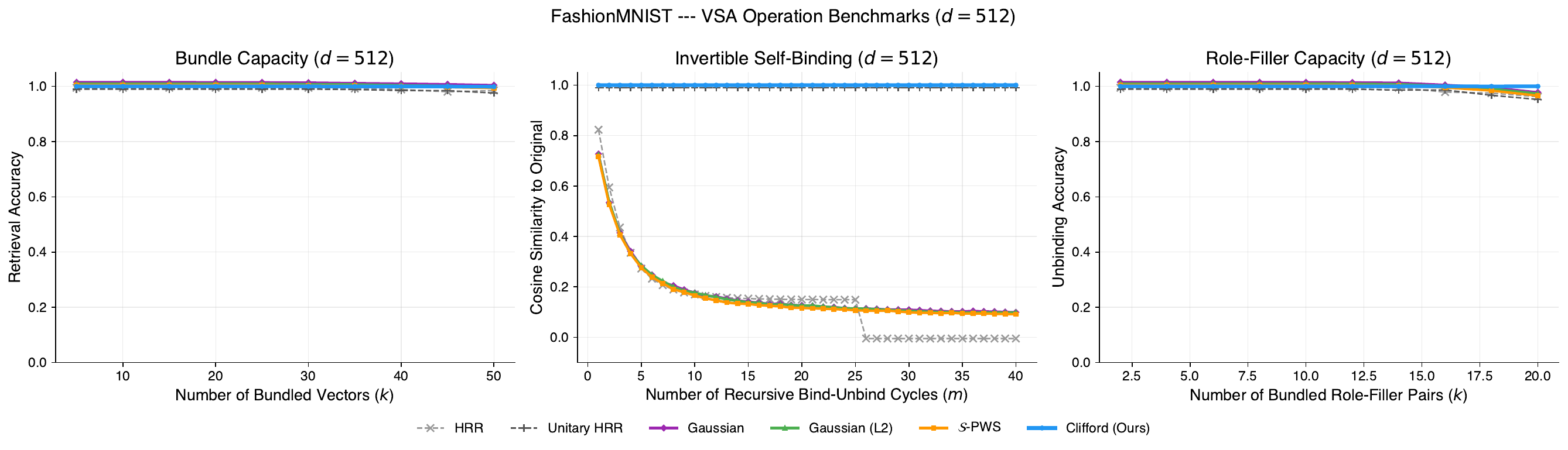}
    \caption{FashionMNIST $d=512$.}
  \end{figure} 
  \begin{figure}[H]
    \centering
    \includegraphics[width=\linewidth]{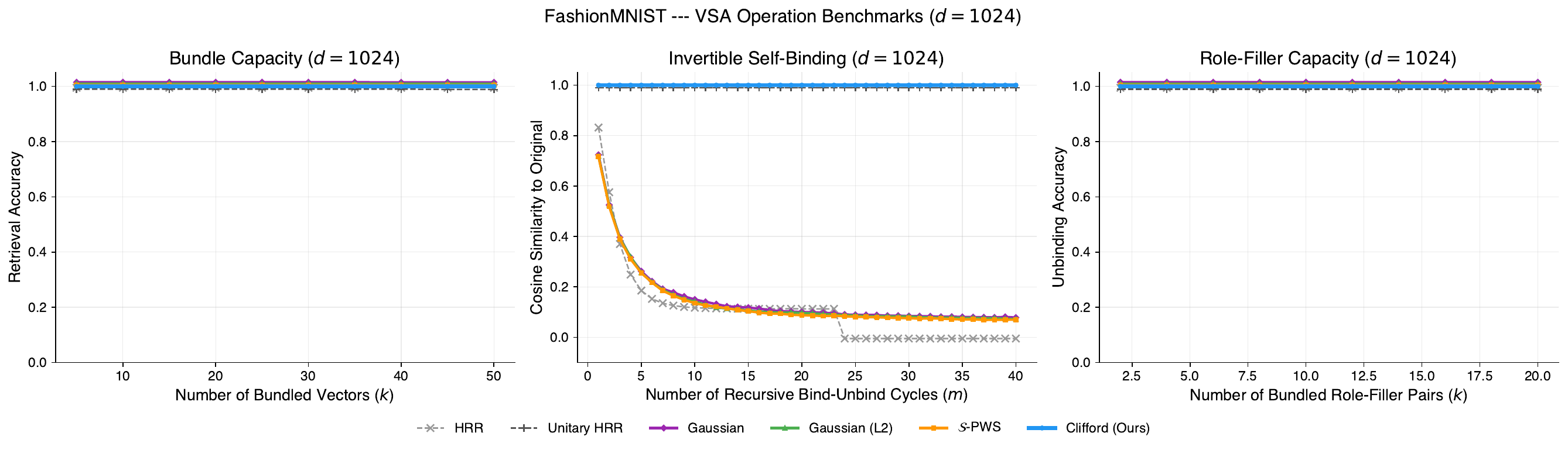}
    \caption{FashionMNIST $d=1024$.}
  \end{figure}
  \begin{figure}[H]
    \centering
    \includegraphics[width=\linewidth]{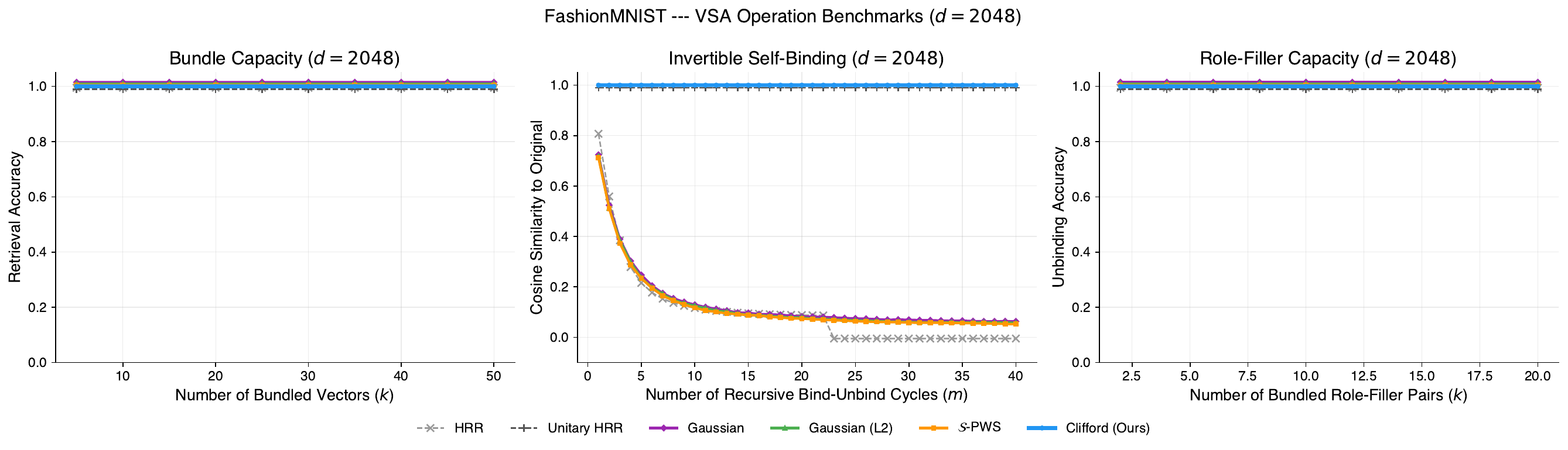}
    \caption{FashionMNIST $d=2048$.}
  \end{figure}
  \begin{figure}[H]   
    \centering
    \includegraphics[width=\linewidth]{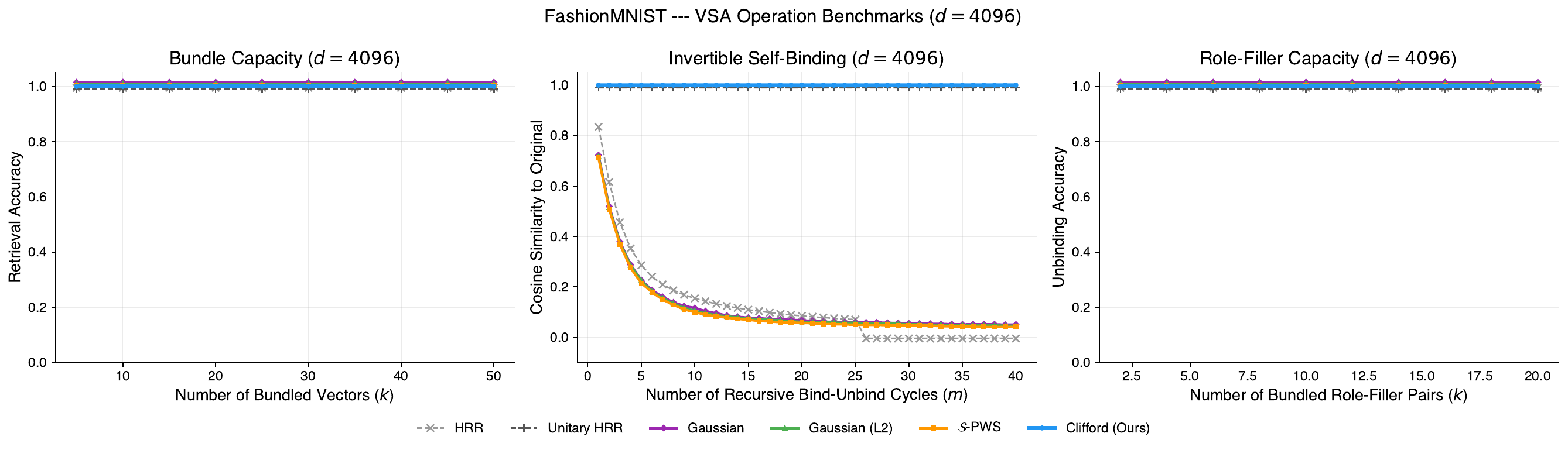}
    \caption{FashionMNIST $d=4096$.}
  \end{figure} 

  \subsection{CIFAR-10 VSA benchmarks}
  \begin{figure}[H]
    \centering
    \includegraphics[width=\linewidth]{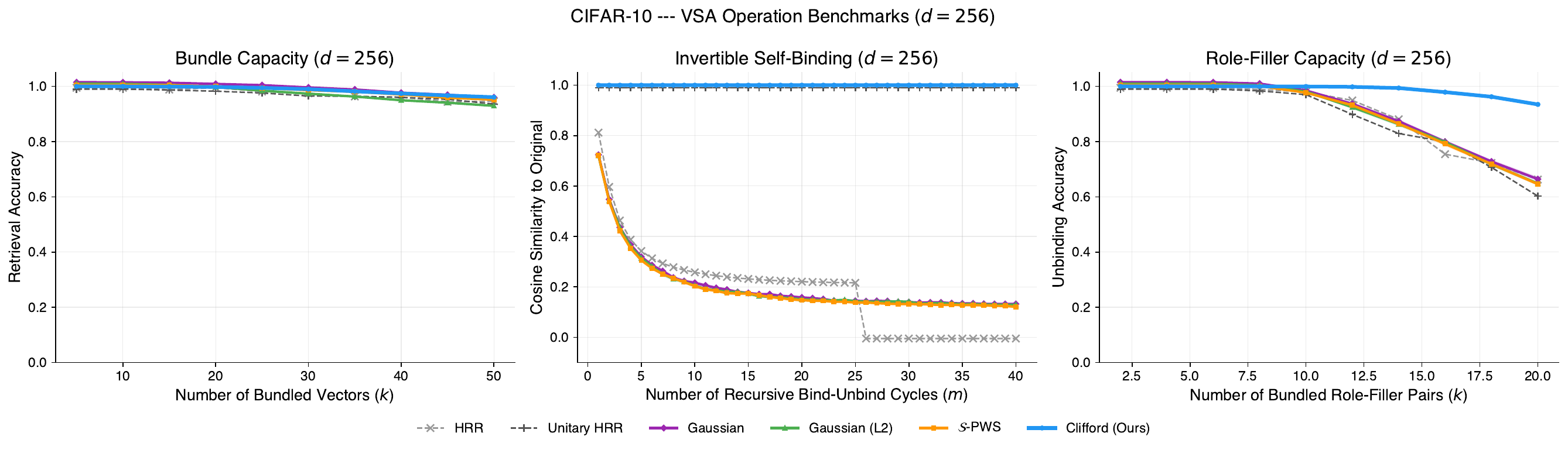}
    \caption{CIFAR-10 $d=256$.}
  \end{figure}
  \begin{figure}[H]
    \centering
    \includegraphics[width=\linewidth]{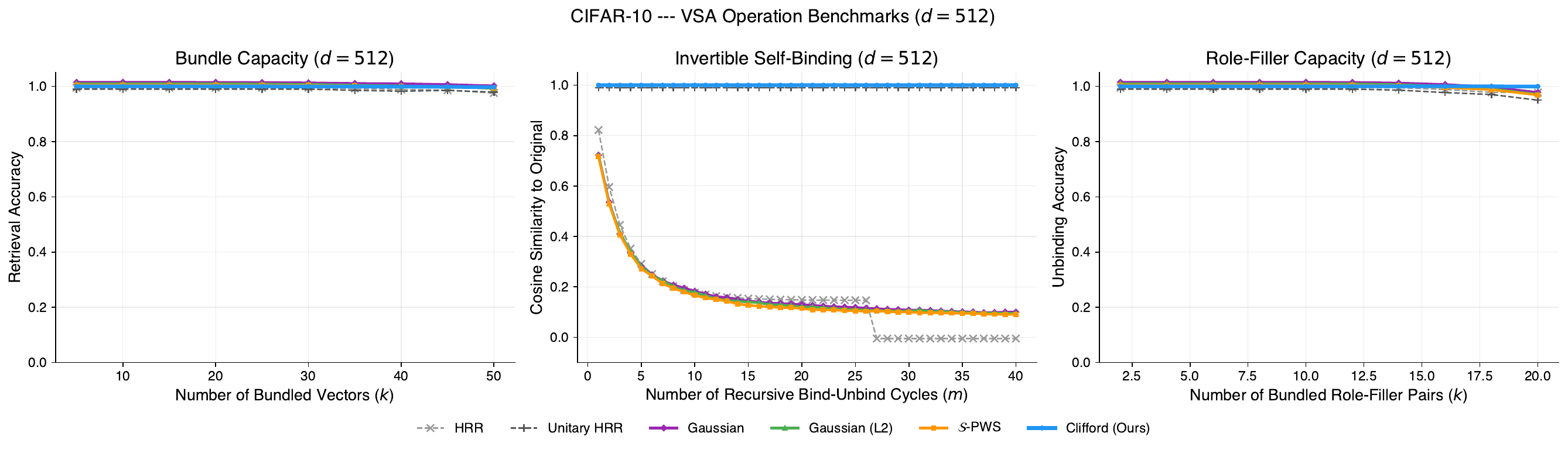}
    \caption{CIFAR-10 $d=512$.}
  \end{figure} 
  \begin{figure}[H]
    \centering
    \includegraphics[width=\linewidth]{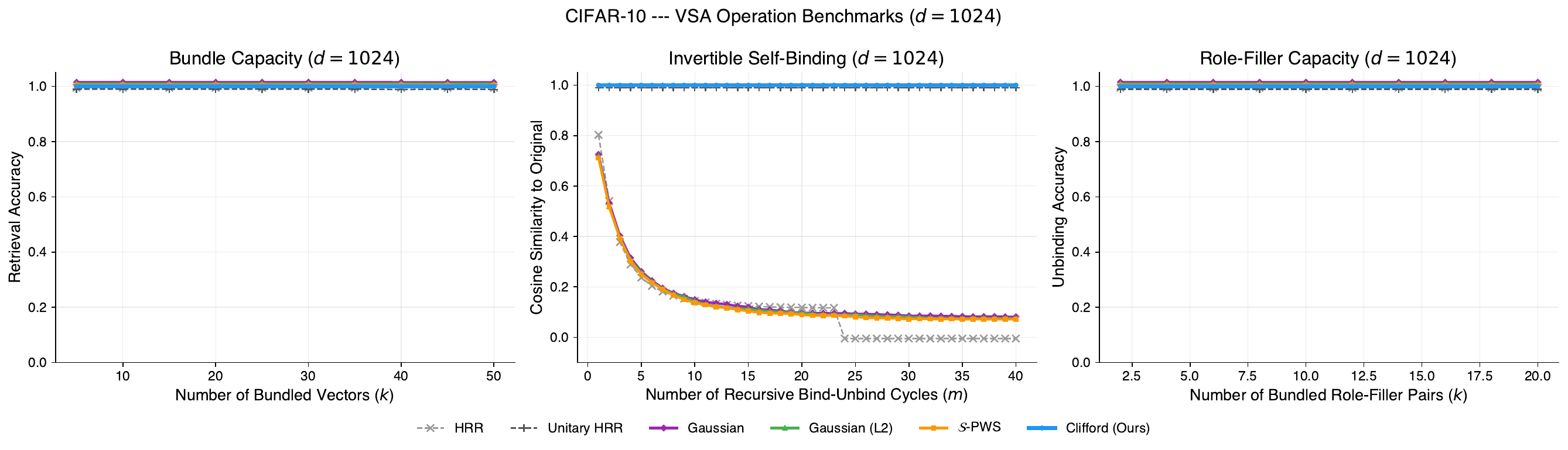}
    \caption{CIFAR-10 $d=1024$.}
  \end{figure}
  \begin{figure}[H]
    \centering
    \includegraphics[width=\linewidth]{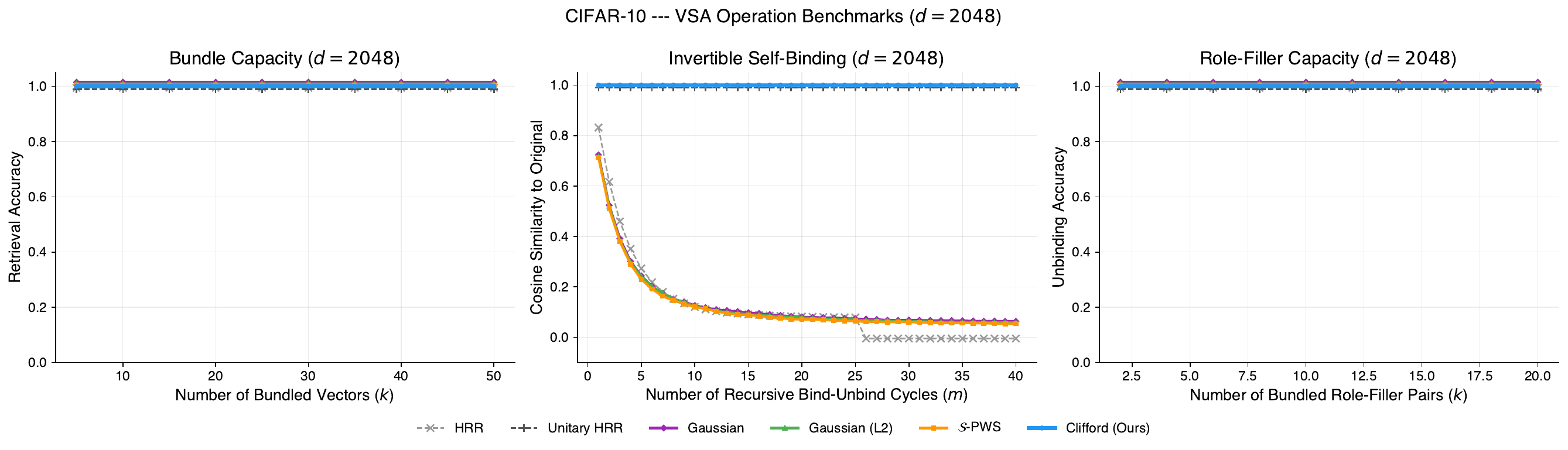}
    \caption{CIFAR-10 $d=2048$.}
  \end{figure}
  \begin{figure}[H]
    \centering
    \includegraphics[width=\linewidth]{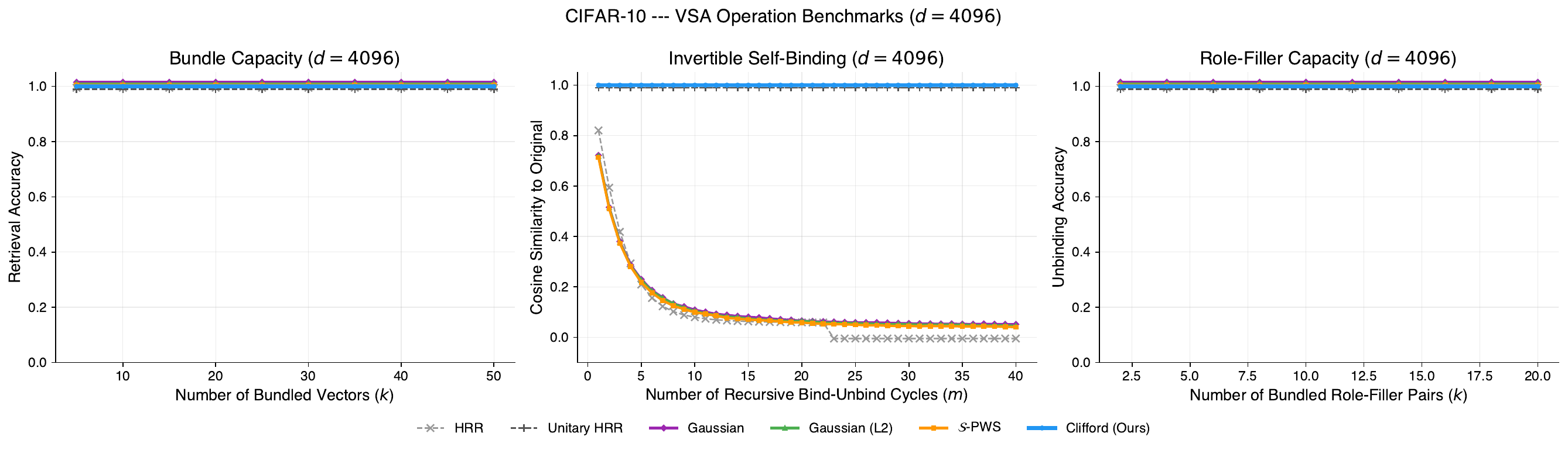}
    \caption{CIFAR-10 $d=4096$.}
  \end{figure} 

  \subsection{MNIST}
  \begin{figure}[H]
    \centering
    \begin{subfigure}[b]{\linewidth}
      \includegraphics[width=\linewidth]{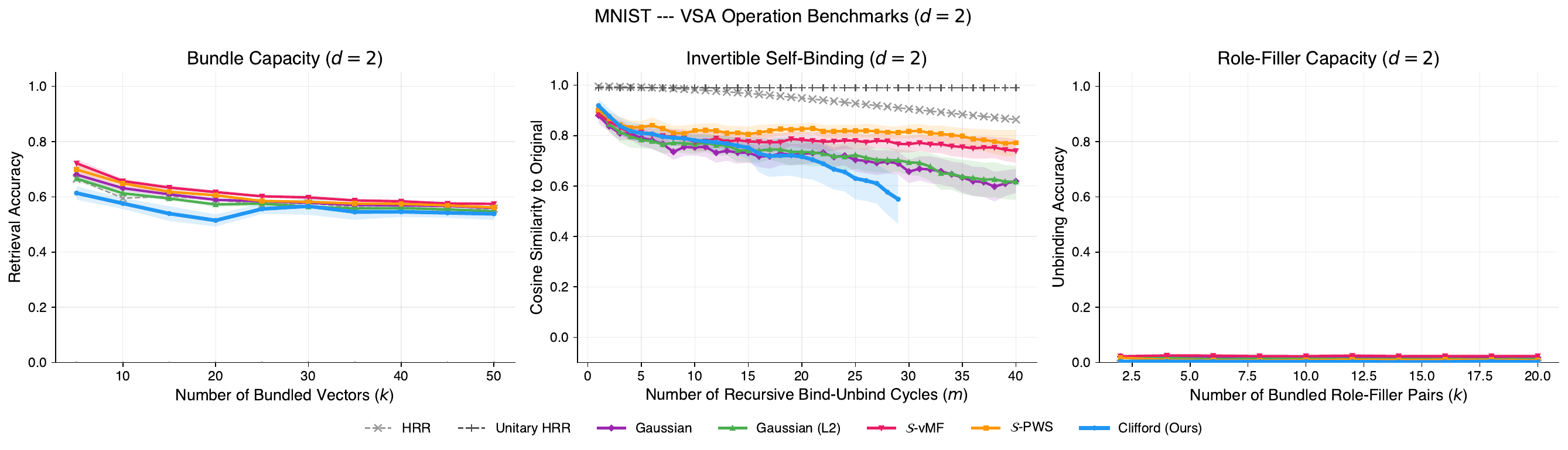}
      \caption{$d=2$.}
    \end{subfigure}\\[4pt]
    \begin{subfigure}[b]{\linewidth}
      \includegraphics[width=\linewidth]{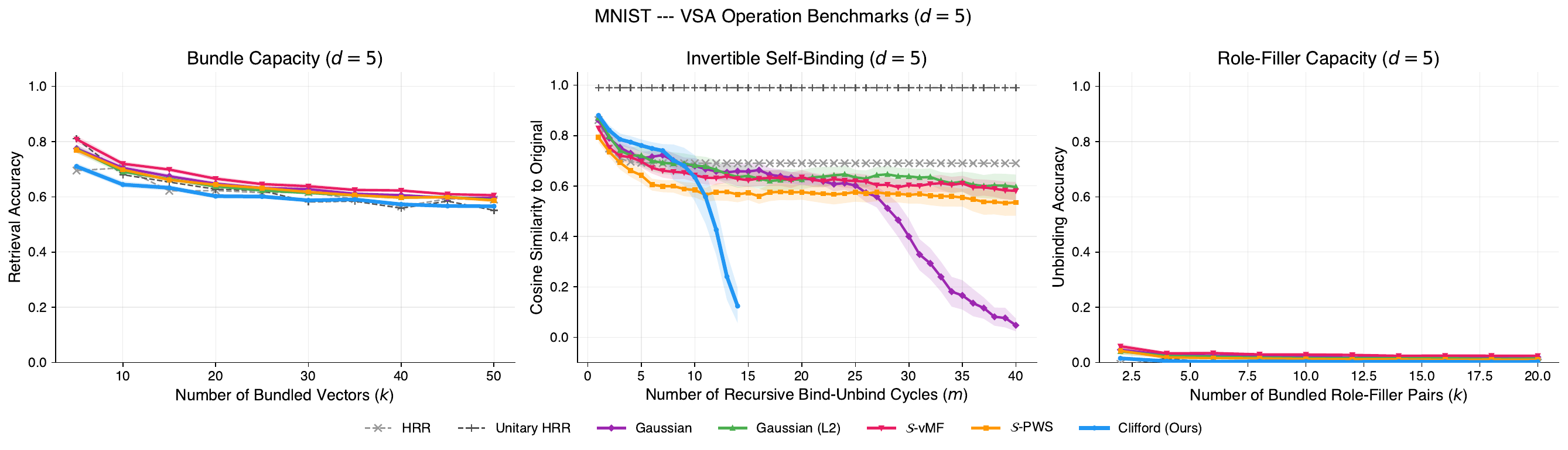}
      \caption{$d=5$.}
    \end{subfigure}\\[4pt]
    \begin{subfigure}[b]{\linewidth}
      \includegraphics[width=\linewidth]{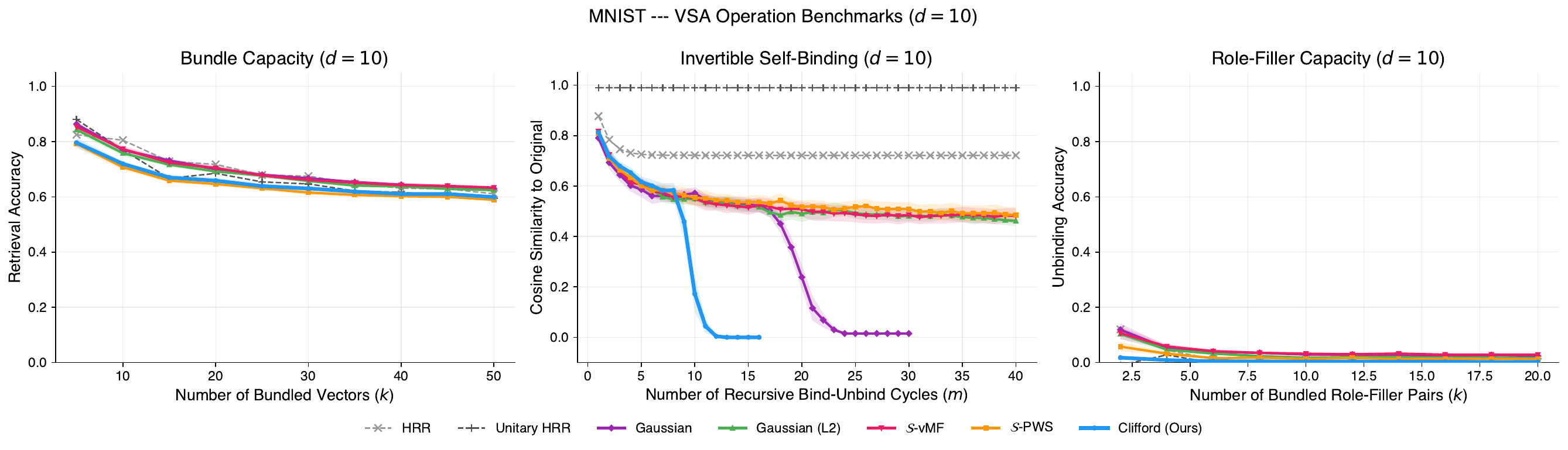}
      \caption{$d=10$.}
    \end{subfigure}   
    \caption{MNIST VSA benchmarks, $d \in \{2,5,10\}$.}
    \label{fig:mnist_vsa_low}
  \end{figure}

  \begin{figure}[H]
    \centering
    \begin{subfigure}[b]{\linewidth}
      \includegraphics[width=\linewidth]{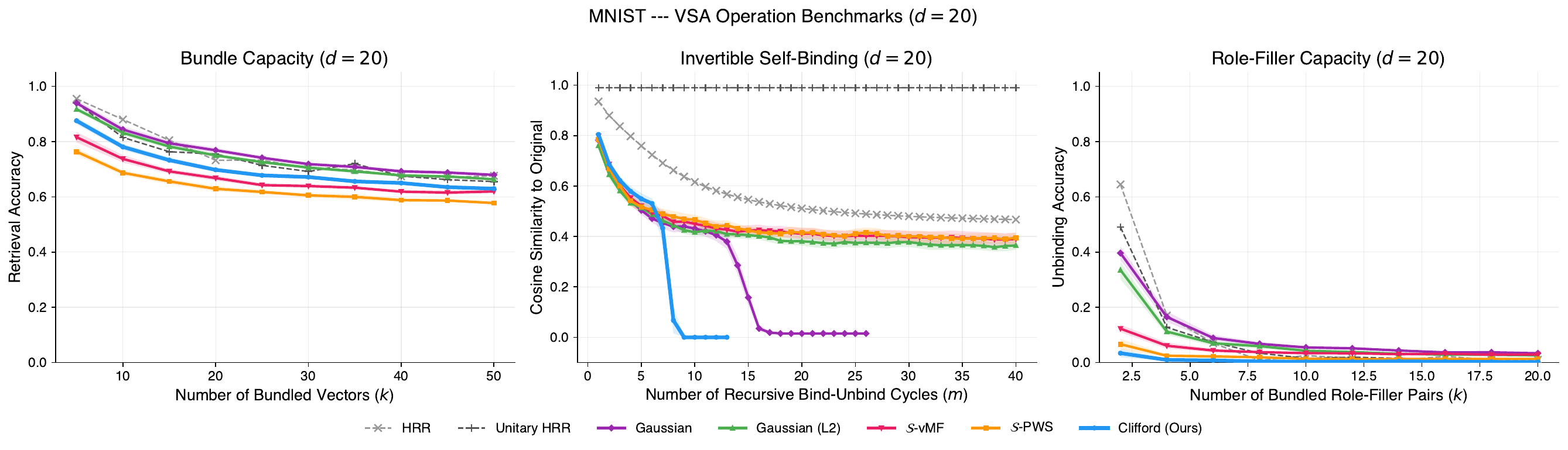}
      \caption{$d=20$.}
    \end{subfigure}\\[4pt]
    \begin{subfigure}[b]{\linewidth}
      \includegraphics[width=\linewidth]{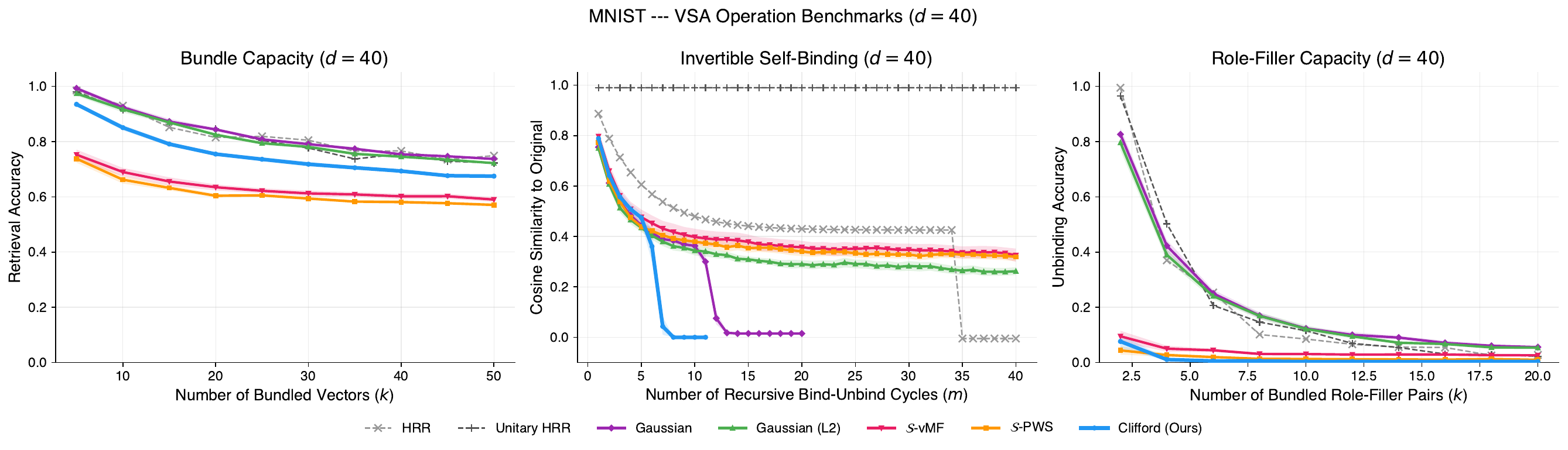}
      \caption{$d=40$.}
    \end{subfigure}   
    \caption{MNIST VSA benchmarks, $d \in \{20,40\}$.}
    \label{fig:mnist_vsa_high}
  \end{figure}

\end{document}